\documentclass{article}

\usepackage{microtype}
\usepackage{graphicx}
\usepackage{subcaption}
\usepackage{booktabs} %

\usepackage{hyperref}

\usepackage[accepted]{icml2024}

\usepackage{amsmath}
\usepackage{amssymb}
\usepackage{mathtools}
\usepackage{amsthm}

\usepackage[capitalize]{cleveref}
\crefname{section}{\S}{\S\S}
\theoremstyle{plain}

\theoremstyle{definition}

\theoremstyle{remark}

\usepackage[textsize=tiny]{todonotes}

\usepackage{algorithm}
\usepackage{algorithmic}

\renewcommand{\algorithmiccomment}[1]{// #1}

\usepackage{amsmath,amsfonts,bm}

\def\eqref#1{equation~\ref{#1}}

\def\1{\bm{1}}

\def\rvp{{\mathbf{p}}}
\def\rvq{{\mathbf{q}}}

\def\vp{{\bm{p}}}
\def\vq{{\bm{q}}}

\def\evp{{p}}
\def\evq{{q}}

\DeclareMathAlphabet{\mathsfit}{\encodingdefault}{\sfdefault}{m}{sl}
\SetMathAlphabet{\mathsfit}{bold}{\encodingdefault}{\sfdefault}{bx}{n}

\def\sS{{\mathbb{S}}}
\def\sT{{\mathbb{T}}}

\DeclareMathOperator*{\argmax}{arg\,max}

\usepackage{hyperref}
\usepackage{url}
\usepackage[acronym]{glossaries}
\glsdisablehyper
\usepackage{todonotes}
\usepackage{amsmath,amsfonts,amssymb}
\usepackage{paralist}
\usepackage{multirow}
\usepackage{enumitem}
\usepackage{makecell}
\usepackage{wrapfig}

\newif\ifsubmission
\submissiontrue

\newif\ifreport
\reporttrue

\ifsubmission
\newcommand{\ahmed}[1]{}
\newcommand{\mcnote}[1]{}
\newcommand{\samnote}[1]{} 
\newcommand{\mjnote}[1]{}
\else
\newcommand{\ahmed}[1]{\todo[color=red!40,inline]{Ahmed: #1}}
\newcommand{\mcnote}[1]{\todo[color=orange!40,inline]{marco: #1}} 
\newcommand{\samnote}[1]{\todo[color=green!40,inline]{Sam: #1}} 
\newcommand{\mjnote}[1]{\todo[color=cyan!40,inline]{MJ: #1}}
\fi

\icmltitlerunning{Flashback: Understanding and Mitigating Forgetting in Federated Learning}

\begin{document}

\twocolumn[
\icmltitle{Flashback: Understanding and Mitigating \\ Forgetting in Federated Learning}

\icmlsetsymbol{equal}{*}

\begin{icmlauthorlist}
\icmlauthor{Mohammed Aljahdali}{yyy}
\icmlauthor{Ahmed M. Abdelmoniem}{comp}
\icmlauthor{Marco Canini}{yyy}
\icmlauthor{Samuel Horváth}{sch}

\end{icmlauthorlist}

\icmlaffiliation{yyy}{KAUST}
\icmlaffiliation{comp}{Queen Mary University of London}
\icmlaffiliation{sch}{MBZUAI}

\icmlcorrespondingauthor{Mohammed Aljahdali}{mohammed.kaljahdali@kaust.edu.sa}

\icmlkeywords{Machine Learning, ICML}

\vskip 0.3in
]

\printAffiliationsAndNotice{}  %

\newacronym{gan}{GAN}{Generative Adversarial Network}
\newacronym{fl}{FL}{Federated Learning}
\newacronym{cl}{CL}{Continual Learning}
\newacronym{nn}{DNN}{Deep Neural Network}
\newacronym{dml}{DML}{Deep Mutual Learning}
\newacronym{kd}{KD}{Knowledge Distillation}
\newacronym{one}{ONE}{On-the-fly Native Ensemble}
\newacronym{dfkd}{DFKD}{Data-Free Knowledge Distillation} %
\newacronym{od}{OD}{Online Distillation}
\newacronym{cifar10}{CIFAR10}{CIFAR10}
\newacronym{iid}{IID}{independent and identically distributed}
\newacronym{kt}{KT}{Knowledge Transfer}
\newacronym{ml}{ML}{Machine Learning}
\newacronym{kt-fl}{KT in FL}{Knowledge Transfer in Federated Learning}
\newacronym{tl}{TL}{Transfer Learning}
\newacronym{cnn}{CNN}{Convolutional Neural Network}
\newacronym{cdf}{CDF}{Cumulative distribution function}

\newacronym{fd}{FD}{Federated Distillation}
\newacronym{fml}{FML}{Federated Mutual Learning}
\newacronym{defkt}{Def-KT}{Decentralized Federated Knowledge Transfer}
\newacronym{fedmd}{FedMD}{Federated Model Distillation}
\newacronym{feddf}{FedDF}{Federated Distillation Fusion}
\newacronym{fedgkt}{FedGKT}{Federated Group Knowledge Transfer}
\newacronym{fedpcl}{FedPCL}{Federated Prototype Contrastive Learning}
\newacronym{fedntd}{FedNTD}{Federated Not-True Distillation}
\newacronym{ktpfl}{KT-pFL}{KT-pFL}
\newacronym{moon}{MOON}{Model-Contrastive Federated Learning}
\newacronym{fedgen}{FedGen}{FedGen}
\newacronym{fedftg}{FedFTG}{FedFTG}
\newacronym{fccl}{FCCL}{Federated Cross-Correlation Learning}
\newacronym{fedavg}{FedAvg}{Federated Averaging}

\newacronym{fedmmd}{FedMMD}{FedMMD}

\graphicspath{{forgetting/}}

\begin{abstract}
    In Federated Learning (FL), forgetting, or the loss of knowledge across rounds, hampers algorithm convergence, particularly in the presence of severe data heterogeneity among clients.
    This study explores the nuances of this issue, emphasizing the critical role of forgetting in FL's inefficient learning within heterogeneous data contexts. Knowledge loss occurs in both client-local updates and server-side aggregation steps; addressing one without the other fails to mitigate forgetting. We introduce a metric to measure forgetting granularly, ensuring distinct recognition amid new knowledge acquisition.
    Leveraging these insights, we propose Flashback, an FL algorithm with a dynamic distillation approach that is used to regularize the local models, and effectively aggregate their knowledge.
    Across different benchmarks, Flashback outperforms other methods, mitigates forgetting, and achieves faster round-to-target-accuracy, by converging in 6 to 16 rounds.
\end{abstract}

\section{Introduction}
\label{sec:intro}

\gls{fl} is a distributed learning paradigm that allows training over decentralized private data. These datasets belong to different clients that participate in training a global model.
\gls{fedavg} \citep{FedAvg-McMahan2017-mv} is a prominent training algorithm that uses a centralized server to orchestrate the process. At every round, the server samples a proportion of the available clients. Starting form the current version of the global model, each sampled client performs $E$ epochs of local training using their private data and sends its updated model to the server. Then, the server aggregates the models by averaging them to obtain the new global model. This process is typically repeated for many rounds until a desired model performance is obtained.

A main challenge in \gls{fl} is the heterogeneity in distribution between the private datasets, which are unbalanced and non-IID \citep{fl-open-problems@Kairouz2019-qi}.
Data heterogeneity causes local model updates to drift -- the local optima might not be consistent with the global optima -- and can lead to slow convergence of the global model -- where more rounds of communication and local computation are needed -- or worse, when the desired performance may not be reached.
Addressing data heterogeneity in \gls{fl} has been the focus of several prior studies. For instance, FedProx~\citep{fedprox@li2020federated} proposes a proximal term to limit the distance between the global model and the local model updates, mitigating the drift in the local updates. MOON~\citep{moon@li2021model} mitigates the local drift using a contrastive loss to minimize the distance between the feature representation of the global model and the local model updates while maximizing the distance between the current model updates and the previous model updates. FedDF~\citep{feddf@lin2020ensemble} addresses heterogeneity in local models by using ensemble distillation during the aggregation step (instead of averaging the model updates).

\begin{figure*}[t!]
\begin{subfigure}{\columnwidth}
    \includegraphics[width=\columnwidth]{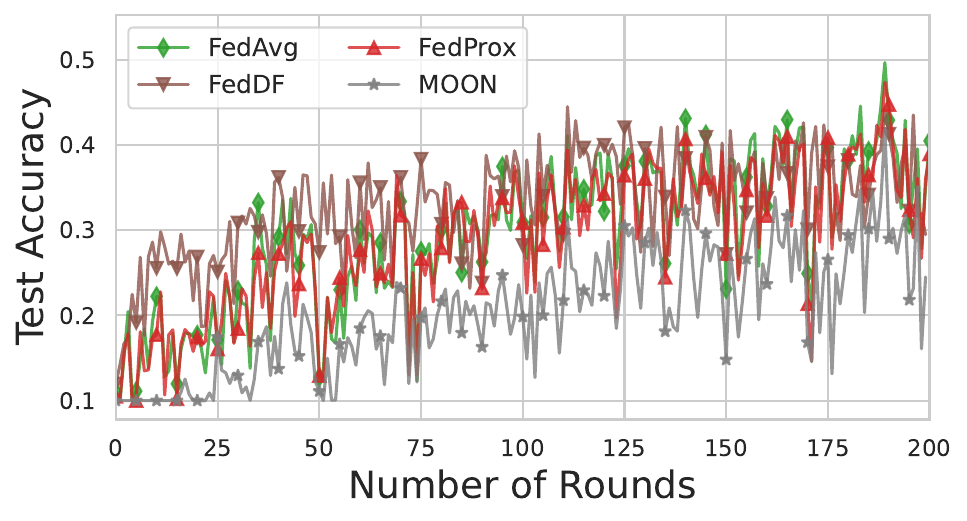}
    \caption{Global model accuracy of \gls{fedavg} and baselines.}
    \label{fig:CIFARDataModule_baselines_test_acc}
\end{subfigure}
\hfill
\begin{subfigure}{\columnwidth}
    \includegraphics[width=\columnwidth]{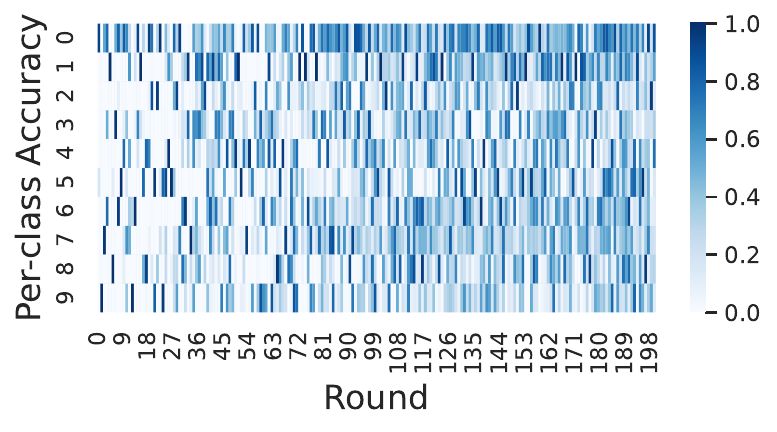}
    \caption{Per-class accuracy of \gls{fedavg}'s global model.}
    \label{fig:CIFARDataModule_FedAvg_Global_Model}
\end{subfigure}
\caption{Performance of \gls{fedavg} and other baselines over training rounds with CIFAR10.}
\label{fig:forgetting}
\end{figure*}

However, we experimentally observe that under severe data heterogeneity, these proposals provide little or even
no advantage over \gls{fedavg}. \Cref{fig:CIFARDataModule_baselines_test_acc} illustrates the test accuracy of \gls{fedavg} and other baselines while training a DNN over the CIFAR10 dataset \citep{cifar10@krizhevsky2009learning} (details in \cref{sec:eval}).

This motivates us to better understand how data heterogeneity poses a challenge for \gls{fl} and devise a new approach of handling non-IID datasets.
We investigate the evolution of the global model accuracy broken down by its per-class accuracy. \Cref{fig:CIFARDataModule_FedAvg_Global_Model} shows as a heatmap the per-class accuracy for \gls{fedavg}; each rectangle represents the accuracy of the global model on a class at a round.
Other baseline methods show similar results.
Our key observation is that there is a significant presence of \emph{forgetting}: i.e., cases where some knowledge obtained by the global model at round $t$ is dropped at round $t+1$, causing the accuracy to decline (as shown by the prominent number of light-shaded rectangles at the right side of darker ones in the figure). 

A similar phenomenon is known as \emph{catastrophic forgetting} in \gls{cl} literature~\citep{forgetting_in_cl@parisi2019continual}.
\gls{cl} addresses the challenge of sequentially training a model on a series of tasks, denoted as $\{T_1, T_2, \ldots, T_n\}$, without revisiting data from prior tasks. Formally, given a model with parameters $\theta$ and task-specific loss functions $ L_t(\theta) $ for each task $ T_t $, the objective in \gls{cl} is to update $ \theta $ such that performance on the current task is optimized without significantly degrading the model's performance on previously learned tasks. This is non-trivial, as na\"ive sequential training often leads to catastrophic forgetting, where knowledge from prior tasks is overridden when learning a new task. An inherent assumption in this paradigm is that once the model transitions from task $ T_i $ to task $ T_{i+1} $, data from $ T_i $ becomes inaccessible, amplifying the importance of knowledge retention strategies~\citep{f_cl2021continual}.

While the premises and assumptions of \gls{fl} differ from those of traditional machine learning and continual learning, forgetting remains an issue. This can be viewed as a side effect of data heterogeneity, a commonality \gls{fl} shares with \gls{cl}. In \gls{fl}, the global model evolves based on a fluctuating data distribution. Specifically, in each communication round, a diverse set of sampled clients, each with distinct data distributions, contribute with a model update.
Furthermore, these model updates need to be aggregated to obtain a global model.
This situation presents dual levels of data heterogeneity. Firstly, at the \textit{intra-round} level, heterogeneity arises from the participation of clients with varied data distributions within the same round. This diversity can inadvertently lead to ``forgetting'' specific data patterns or insights from certain clients. Secondly, at the \textit{inter-round} level, the participating clients generally change from one round to the next. As a result, the global model may ``forget'' or dilute insights gained from clients in previous rounds.

To remedy this issue, we propose Flashback, a \gls{fl} algorithm that employs a dynamic distillation approach to mitigate the effects of data heterogeneity. 
Flashback's dynamic distillation ensures that the local models learn new knowledge while retaining knowledge from the global model during the client updates by adaptively adjusting the distillation loss. Moreover, during the server update, Flashback uses a very small public dataset as a medium to integrate the knowledge from the local models to the global model using the same dynamic distillation. 
Flashback performs these adaptations by estimating the knowledge in each model using label counts as a proxy of the model knowledge.
Overall, Flashback results in a more stable and faster convergence compared to existing methods.

Our contributions are the following:\\
$\bullet$ We investigate the forgetting problem in FL. We show that under severe data heterogeneity, \gls{fl} sufferers from forgetting. We dissect how and where forgetting happens (\cref{sec:forgetting}).\\
$\bullet$ We propose a new metric for measuring forgetting over the communication rounds(\cref{sec:forgetting}).\\
$\bullet$ We introduce \emph{Flashback}, a \gls{fl} algorithm that employs a dynamic distillation during the local updates and the server update (\cref{sec:flashback}). By addressing the forgetting issue, Flashback not only mitigates its detrimental effects but also converges to the desired accuracy faster than existing methods (\cref{sec:eval}).

\section{Background}

We consider a standard cross-device \gls{fl} setup in which there are $N$ clients. Each client $i$ has a unique dataset $D_i = \{(x_j, y_j)\}_{j=1}^{n_i}$ where $x_j$ represents the input features and $y_j$ is the ground-truth label for $j$-th data point and $n_i$ represent the size of the local dataset of client $i$. The goal is to train a single global model that minimizes the objective:
\begin{equation*}
   \min_{w \in \mathbb{R}^d} \sum^N_{i=1} \frac{|D_i|}{|\cup_{i \in [N]} D_i|} \left\{L_i(w) =  \frac{1}{|D_i|}\sum_{j=1}^{|D_i|} l(w; (x_j, y_j))\right\},
\end{equation*}
where $L_i(w)$ represents the local loss for client $i$, and $l(w; (x_j, y_j)) = \mathcal{L}_{\text{CE}}(F_{w}(x), y)$ is the cross-entropy loss for a single data point,
where $F_{w}$ denotes the model parameterized by learnable weights $w$.

\gls{fedavg} provides a structured approach to efficiently address this distributed problem. At each communication round $t$, the server randomly selects $K$ clients from the total available $N$ clients. These clients (denoted with $\sS_t$) receive the previous global model, $w_{t-1}$. Then, they update this model based on their local data using their local loss function $L_i$. After updating, each client sends their modified model $w_{k,t}$ back to the server that updates the global model using a weighted average of local models, i.e., 
$ w_t = \sum_{k \in \sS_t} \frac{|D_k|w_{k,t}}{|\cup_{k \in [K]} D_k|}.$
To accommodate the intrinsic heterogeneity in client data, various \gls{fl} algorithms introduce modifications either at the local update level or during the global aggregation. The nuances of these variations are further explored in \cref{sec:rw}. 

\begin{figure}[t!]
    \centering
    \includegraphics[width=\linewidth]{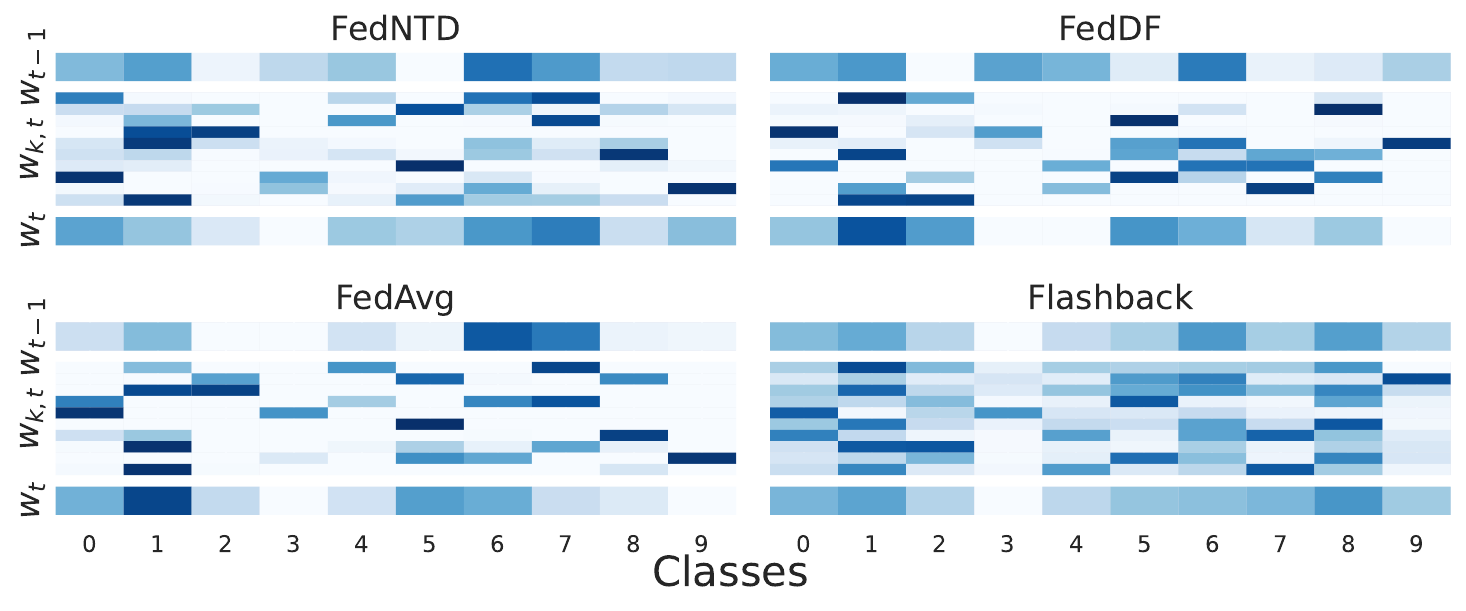}
    \vspace{-1.5em}
    \caption{Local (client) \& Global Forgetting in some of the baselines using CIFAR10. The first row represents the global model per-class test accuracy at round $t-1$; then, the rows in the middle are all the clients that participated in round $t$, and finally, in the last row, the global model at the end of round $t$. 
    Local forgetting happens when clients at round $t$ lose the knowledge that the global model had at round $t-1$. The global forgetting happens when the global model at round $t$ loses the knowledge that in the clients' models at round $t$.}
    \label{fig:global_local_forgetting_cifar_fedavg}
    
\end{figure}

Among these, \textbf{Knowledge Distillation (KD)} is a training method wherein a smaller model, referred to as the student, is trained to reproduce the behavior of a more complex model or ensemble called the teacher. Let $F_{w_s}$ denote the student model with weights $w_s$ and $F_{w_t}$ represent the teacher model with weights $w_t$. For a given input $x$, the student aims to minimize the following distillation loss:
\begin{equation}
\label{eq:distillation}
\begin{aligned}
    \mathcal{L}_{\text{KD}}((x,y); w_s, w_t) = \mathcal{L}_{\text{CE}} (F_{w_s}(x), y) (1-\alpha)  \\ +  \mathcal{L}_{\text{KL}}(F_{w_t}(x), F_{w_s}(x)) \alpha
\end{aligned}
\end{equation}
Here, $\mathcal{L}_{\text{CE}}$ is the standard cross-entropy loss with true label $y$, and $\mathcal{L}_{\text{KL}}$ represents the Kullback-Leibler (KL) divergence between the teacher's and the student's output probabilities. It is defined as $\mathcal{L}_{\text{KL}}(\vp, \vq) = \sum_{c=1}^C \evp^c \log \left( \frac{\evp^c}{\evq^c} \right)$, where $C$ is the number of classes, $\rvp$ is the target output probability vector, and $\rvq$ is the predicted output probability vector.
The hyperparameter $\alpha \in [0, 1]$ balances the importance between the learning from the true labels and the teacher's outputs.

While distillation originally emerged as a method for model compression~\cite{hinton2015distilling,bucilua2006model,schmidhuber1991neural}, its utility extends to \gls{fl}. In the federated context, distillation can combat challenges like data heterogeneity ~\cite{feddf@lin2020ensemble, FedNTD-Lee2021-is} and communication efficiency ~\cite{FD-Jeong2018-ew}.
Specifically, the global model can act as a guiding teacher during local updates, directing the training process for each client. Additionally, distillation techniques streamline the aggregation step, assisting in the incorporation of varied knowledge from diverse clients to update the global model. Furthermore, employing distillation for aggregation mitigates model heterogeneity, allowing for the use of different model architectures. Perhaps most interestingly, by using distillation, \gls{fl} systems can potentially bypass the traditional method of transmitting weight updates. This is accomplished by sending soft labels that encapsulate the essence of local updates--communication becomes more efficient, reducing bandwidth usage.

\section{Forgetting in FL}
\label{sec:forgetting}

We now investigate where forgetting happens and devise a metric to quantify this phenomenon.
Recall that in \gls{fl}, the models are updated in two distinct phases: 1) during local training -- when each client $k$ starts from global model $w_{t-1}$ and locally trains $w_{k, t}$ -- and 2) during the aggregation step -- when the server combines the client models to update the new global model $w_{t}$.

Intuitively, forgetting in \gls{fl} is when knowledge contained in the global model will be lost after the completion of communication round $w_{t-1} \rightarrow w_{t}$.
We observe that forgetting may occur in the two phases of \gls{fl}.
We refer to the former case as \textbf{local forgetting}, where some knowledge in the global model will be lost during the local training $w_{t-1} \rightarrow w_{k,t}$. This is due to optimizing for the clients' local objectives, which depend on their datasets. Local forgetting is akin to the form of forgetting seen in \gls{cl}, where tasks change over time (as with clients in \gls{fl}) and, consequently, the data distribution.
We refer to the latter case as \textbf{aggregation forgetting}, where some knowledge contained in the clients' model updates will be lost during aggregation $\sum \{w_{k,t} \mid k \in \sS_t \} \rightarrow w_{t}$. This might be due to the coordinate-wise aggregation of weights as opposed to matched averaging in the parameter space of DNNs~\citep{fedma@wang2020federated}.

We illustrate forgetting in \cref{fig:global_local_forgetting_cifar_fedavg} based on actual experiments with several baseline methods. The figure shows the per-class accuracy of $w_{t-1}$, all local models $w_{k,t}$, and the new global model $w_{t}$. The local forgetting is evident in the drop in accuracy (lighter shade of blue) of the local models $w_{k,t}$ compared to the global model $w_{t-1}$. The aggregation forgetting is evident in the drop in accuracy of the global model $w_{t}$ compared to the local models $w_{k,t}$. The figure also previews a result of our method, Flashback, which shows a significant mitigation of forgetting. Local and aggregation forgetting lead to the main forgetting problem in FL, which we term as \textbf{round forgetting}, affecting $w_{t-1} \rightarrow w_{t}$.

In \gls{cl}, forgetting is often quantified using Backward Transfer (BwT)~\citep{forgetting_metric@chaudhry2018riemannian}.
\citet{FedNTD-Lee2021-is} adapted this metric for \gls{fl} as:
\begin{equation}
\label{eq:forgetting_score}
\textstyle    \mathcal{F} = \frac{1}{C} \sum_{c=1}^C \argmax_{t\in{1, T-1}} (A^c_t - A^c_T),
\end{equation}
where $C$ is the number of classes and $A^c_t$ is the global model accuracy on class $c$ at round $t$.

However, $\mathcal{F}$ is a coarse-grain score that evaluates forgetting in aggregate across all rounds.
We seek a finer-grain metric that measures forgetting round-by-round. Furthermore, we wish to account for knowledge replacement scenarios, such as when a decline in accuracy for one class might be accompanied by an increase in another, essentially masking the negative impact of forgetting in aggregate measures.
Thus, for our evaluation results (\cref{sec:eval}), we propose to measure \textbf{round forgetting} by focusing only on drops in accuracy using the following metric:
\begin{equation*}
\textstyle    \mathcal{F}_t = - \frac{1}{C}\sum_{c=1}^C \min (0, (A^c_t - A^c_{t-1}))
\end{equation*}
where $t > 1$ is the round at which forgetting is measured.

Our metric accounts for the pitfalls of the previously proposed forgetting metric. It discounts knowledge replacement scenarios that can happen between rounds by only focusing on the negative changes in accuracy. Furthermore, it provides a granular view of forgetting, because it measures forgetting within the rounds (whereas \cref{eq:forgetting_score} measures how much the global model has forgotten by the end of training).

\ifreport
\ifreport
\subsection{Forgetting in Federated Learning and Continual Learning}
\else
\section{Forgetting in Federated Learning and Continual Learning}
\fi

Forgetting is a prominent problem in \gls{cl}, where tasks change over time, and consequently, the data distribution, places models at risk of overriding previously learned knowledge. Looking at \gls{fl} with the same perspective, we have the intra-round and inter-round data distribution heterogeneity.
Intra-round clients with different data distributions participate, by updating the current global model weights $w_{t-1}$. The server obtains a set of $\{w_{k, t} \mid k \in S_t \}$ where $S_t$ is the set of clients participating in round $t$. The goal is after the aggregation step, a new global model $w_t$ is obtained containing all the knowledge that was present in $\{w_{k, t} \mid k \in S_t \}$. 
Inter-round the global model $w_t$ has learned knowledge that over the prior rounds $1 \ldots t$. At each round, different sets of clients participate. The goal is to carry this knowledge to the next round $t+1$ even though the new set $S_{t+1}$ most likely is different with respect to the data distribution than the previous set $S_t$. 
Presenting distinct forgetting challenges in \gls{fl} compared to \gls{cl}.

\fi

\section{Forgetting-Robust Federated Learning}
\label{sec:flashback}

Our key idea to mitigate forgetting is to leverage a dynamic form of knowledge distillation, which is fine-tuned in response to the evolving knowledge captured by the different models in the training process.
During local training, distillation ensures that each local model learns from the client's local dataset while retaining knowledge from the current global model.
On the server side, after the clients' updates, Flashback begins by aggregating the locally updated models—much in the vein of \gls{fedavg}. Then, Flashback distills the knowledge of the freshly updated global model using our dynamic distillation, learning from both its immediate predecessor---the global model obtained at the previous round---and the ensemble of locally updated models, which all play the role of teachers.
The Flashback algorithm is detailed in~\cref{algo:flashback}.
The remainder of this section discusses our distillation approach in detail.

\begin{algorithm}[t!]
    \caption{Flashback algorithm.}
    \label{algo:flashback}
    \begin{algorithmic}[1]
        \INPUT Initial global model $w_0$, number of rounds $T$, fraction of clients $R$, minibatch size $B$, number of local epochs $E$, number of server epochs $E_s$, learning rate $\eta$
        \OUTPUT global model $w_T$
        
        \STATE $\boldsymbol{\pi} = \boldsymbol{0} \in \mathbb{R}^C$ \COMMENT{Global model's label count vector}
        \FOR{$t = 1$ {\bfseries to} $T$}
            \STATE $\sS_t \leftarrow$ Randomly select $\lceil R \cdot N \rceil$ clients
            \FOR{each client $k \in \sS_t$}
                \STATE $w_{k, t} \leftarrow w_{t-1}$ \COMMENT{Initialize local model with current global model}  
                \STATE $B_k \leftarrow$ Split local dataset into batches of size $B$
                \STATE Compute $\boldsymbol{\alpha}$ with $\boldsymbol{\nu}$ as the local label count and a single teacher $\boldsymbol{\mu} \leftarrow \boldsymbol{\pi}$
                \FOR{$e = 1$ {\bfseries to} $E$}
                    \FOR{each batch $b \in B_k$}
                        \STATE Update $w_{k, t}$ using dKD loss $\mathcal{L}_{\text{dKD}}$
                    \ENDFOR
                \ENDFOR
            \ENDFOR
            \STATE $m_t \leftarrow \sum_{k \in \sS_t} n_k$ \COMMENT{Total data points in this round}
            \STATE $w_{t} \leftarrow \sum_{k \in \sS_t} \frac{n_k}{m_t} w_{k, t}$ \COMMENT{Average to obtain the new global model}
            \STATE $B_s \leftarrow$ Split the public dataset into batches of size $B$
            \STATE $\sT \leftarrow \{ w_{k, t} \mid k \in \sS_t \} \cup \{ w_{t-1}\}$
            \STATE Compute $\boldsymbol{\alpha}$ with $\boldsymbol{\nu} \leftarrow \boldsymbol{\pi}$ and $\boldsymbol{\mu}_i$ as the label count $\forall w_i \in \sT$
            \FOR{$e = 1$ {\bfseries to} $E_s$}
                \FOR{each batch $b \in B_s$}
                    \STATE Update $w_{t}$ using dKD loss $\mathcal{L}_{\text{dKD}}$
                \ENDFOR
            \ENDFOR
            \STATE $r_k \leftarrow$ (Increment $r_k$ for every client $k \in \sS_t$)\;
            \STATE Update participation count for each client $k \in \sS_t$
            \FOR{each client $k \in \sS_t$}
                \IF{$\gamma r_k \leq 1$}
                    \STATE $\boldsymbol{\pi} \leftarrow \boldsymbol{\pi} + \gamma \boldsymbol{\mu}_k$ \label{algo:flashback:pi}
                \ENDIF
            \ENDFOR
        \ENDFOR
    \end{algorithmic}
\end{algorithm}

\subsection{Dynamic Distillation}

As established in~\cref{sec:forgetting}, a client's local model can forget and override model knowledge with what is present in its private data.
Moreover, even the global model can be imperfect for two reasons: 
\begin{inparaenum}[i)]
    \item As we established before, the global model is susceptible to forgetting in the aggregation step.
    \item Assuming no forgetting in the aggregation step, the knowledge contained in the clients who participated so far might not represent all the available knowledge, especially in the early rounds.
\end{inparaenum}
Overall, both local models and the global model can be imperfect. Therefore, the logits of all the different classes cannot be treated equally. 

We propose to use the label count as an approximation of the knowledge within a model.
Here, the label count refers to the occurrences of each class in the training data that the model saw during training.

In machine learning, a model's knowledge is fundamentally tied to the data it has been exposed to. If certain classes have higher representation (or label counts) in the training data, it's intuitive that the model would have more opportunities to learn the distinguishing features of such classes. Conversely, underrepresented classes might not offer the model sufficient exposure to learn their nuances effectively.
Our experimental results suggest that per-class performance on the test set correlates highly with the label counts in the training data. In scenarios where certain classes were more abundant, the model demonstrated higher proficiency in predicting those classes on the test set. As an example, \cref{fig:label_count_vs_test_acc_cinic10_fedavg} in the appendix, illustrates for a randomly-chosen client that the client's model performance on the test set well reflects the label count distribution of its private data. From this and many similar observations, we conclude that the label count can be indicative of a model's knowledge.

In standard knowledge distillation (\cref{eq:distillation}), all logits are treated equally since it is assumed that the teacher model has been trained on a balanced dataset. Owing to the heterogeneity of data distribution in local datasets, this assumption does not hold in \gls{fl}.
As a result, we cannot directly treat the current global model nor the local model updates as equally reliable teachers across all classes.
Instead, we propose weighting the logits using the label count as an approximation of the per-class knowledge within a model.

We now revisit the distillation loss in~\cref{eq:distillation} and transform the scalar $\alpha$ to a matrix form that is automatically tuned according to the label count of both the student and the teachers and used directly within the KL divergence loss.
Namely, the dynamic $\boldsymbol{\alpha}$ parameter (defined below) will change during the training as the label counts change. Flashback maintains the global model counts over the rounds; this mechanism is detailed in the next section.

We consider a single student model $F_{w_s}$ with weights $w_s$ and a set $\sT$ of $K$ teacher models; the $i$-th teacher model is denoted as $F_{w_i}$ with weights $w_i$.
Let $\boldsymbol{\nu} \in \mathbb{R}^{C}$ be the relative label count vector of the student model, where $\nu^c$ is the relative occurrences of class $c$ in the dataset.
Similarly, let $\boldsymbol{\mu}_i \in \mathbb{R}^{C}$ be the relative label count vector of the $i$-th teacher model.

The dynamic $\boldsymbol{\alpha} \in [0, 1]^{K \times C}$ is defined as $[\boldsymbol{\alpha}_1^\intercal, \ldots, \boldsymbol{\alpha}_K^\intercal]$, with $\alpha^c_i = \frac{\mu^c_i}{\nu^c + \sum_k \mu^c_k}$.

Then, we embed $\boldsymbol{\alpha}$ directly in the KL divergence loss ($\mathcal{L}_{\text{KL}}$ in~\cref{eq:distillation}) as follows:
\begin{equation*}
    \mathcal{L}_{\text{dKL}}(\vp, \vq; \boldsymbol{\alpha}_i) = \sum_{c=1}^C \alpha_i^c \cdot \evp^c \log \left( \frac{\evp^c}{\evq^c} \right)
\end{equation*}

Similar to standard distillation, to account for the student model knowledge with respect to the ground-truth class $y$, we define $\alpha^c_s = \frac{\nu^c}{\nu^c + \sum_k \mu^c_k}$. Thus, $\alpha^c_s + \sum_{k=1}^K \alpha^c_k = 1$ for all classes $c \in [C]$.

Finally, the dynamic knowledge distillation loss ($\mathcal{L}_{\text{dKD}}$) is:
\begin{equation}
\begin{aligned}
    \mathcal{L}_{\text{dKD}}((x, y); w_s, \sT, \boldsymbol{\alpha}) = \alpha^y_s \mathcal{L}_{\text{CE}}(F_{w_s}(x), y) \\
    \;+ \sum_{{w_i} \in \sT} \mathcal{L}_{\text{dKL}}(F_{w_i}(x), F_{w_s}(x); \boldsymbol{\alpha}_i)
\end{aligned}
\label{eq:d_kd_loss}
\end{equation}

The dynamic $\boldsymbol{\alpha}$ will weigh the divergence between the logits of different classes, making the student model focus more on learning from the teacher's strengths while being cautious of its weaknesses.
This is of great importance in \gls{fl} because of the data heterogeneity problem.
For instance, in the initial training rounds, the global model may not encounter certain classes. Our distillation approach assigns a zero weight to the divergence of these classes, shielding the client model from adopting unreliable knowledge from the global model. Similarly, if a client possesses significantly larger data for a specific class compared to what the global model has encountered, the weight assigned to that class's divergence will be small. This implies that the client model's remains more grounded in classes where it has more comprehensive data.

A property of our distillation is that it will ignore the global model as a teacher in the first communication round. Since the initial global model label counts are all zeros, \cref{eq:d_kd_loss} reduces to just the cross-entropy: $\mathcal{L}_{\text{dKD}} = \mathcal{L}_{\text{CE}}$.

\begin{figure*}[t!]
\centering
\includegraphics[width=0.5\textwidth]{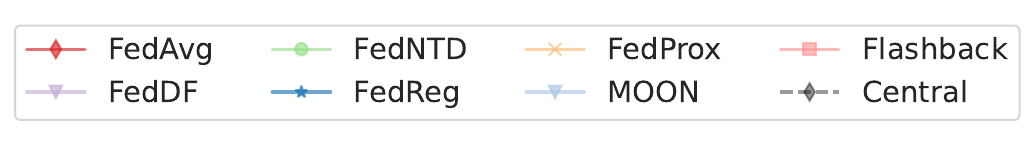}
\vfill
\begin{subfigure}{0.33\textwidth}
    \includegraphics[width=\textwidth]{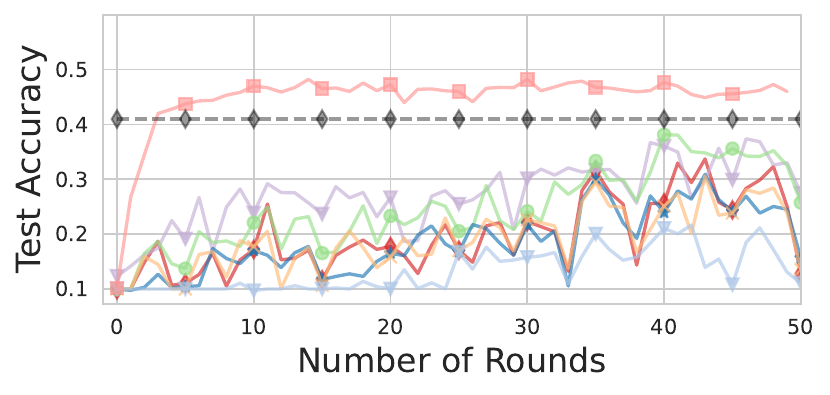}
    \caption{CIFAR10.}
    \label{fig:cifar10_main_test_acc}
\end{subfigure}
\hfill
\begin{subfigure}{0.33\textwidth}
    \includegraphics[width=\textwidth]{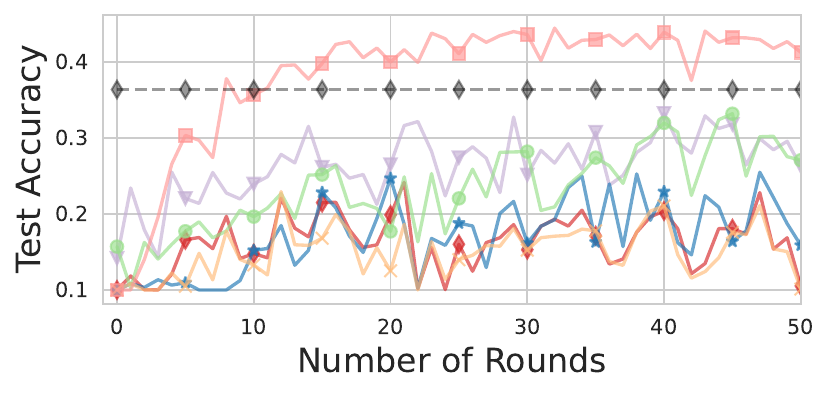}
    \caption{CINIC10.}
    \label{fig:cinic10_main_test_acc}
\end{subfigure}
\hfill
\begin{subfigure}{0.33\textwidth}
    \includegraphics[width=\textwidth]{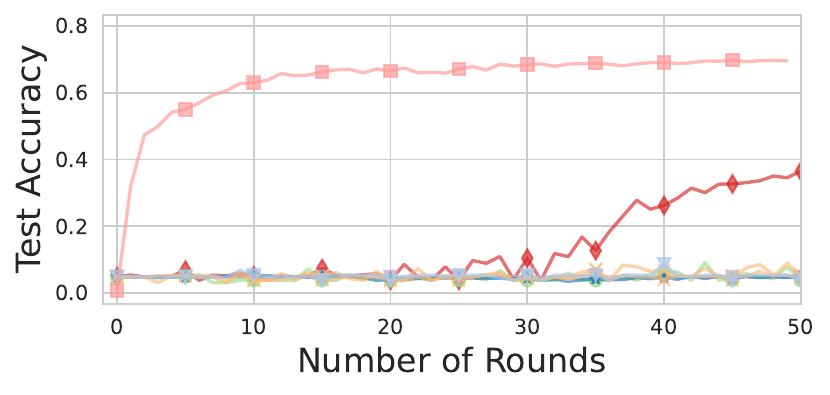}
    \caption{FEMNIST.}
    \label{fig:femnist_main_test_acc}
\end{subfigure}
\vspace{-2.0em}
\caption{Round-to-accuracy performance of Flashback and other baselines over training rounds.}
\label{fig:main_test_acc}
\end{figure*}

\ifreport
\begin{figure*}[t!]
\centering
\includegraphics[width=0.8\textwidth]{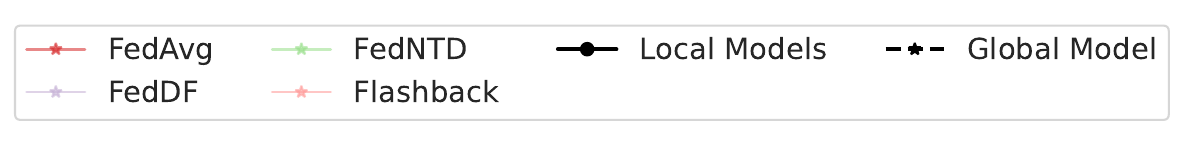}
\begin{subfigure}{0.49\textwidth}
    \includegraphics[width=1\textwidth]{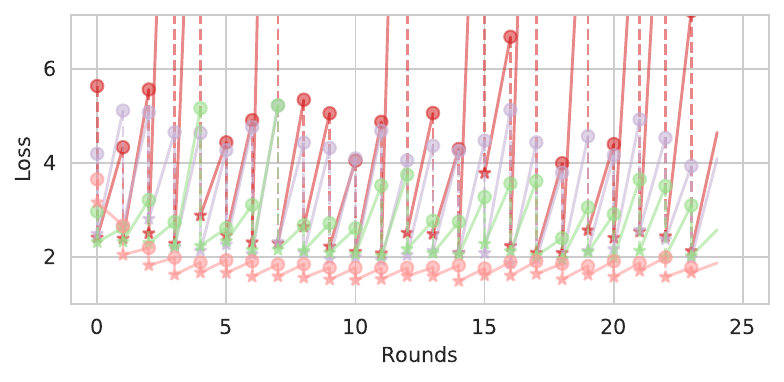}
    \caption{CIFAR10.}
    \label{fig:cifar10_main_main_local_global_test_loss}
\end{subfigure}
 \hfill
\begin{subfigure}{0.49\textwidth}
    \includegraphics[width=1\textwidth]{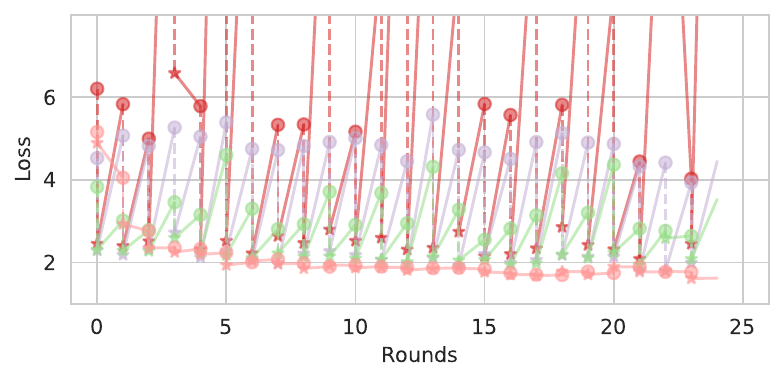}
    \caption{CINIC10.}
    \label{fig:cinic10_main_main_local_global_test_loss}
\end{subfigure}
\vspace{-1.0em}
\caption{Transition of local models loss to the global model loss over the rounds.}
\label{fig:main_main_local_global_test_loss}
\end{figure*}
\fi

\subsection{Estimating the Global Model Knowledge}

Note that to apply the dynamic distillation loss~\cref{eq:d_kd_loss}, we require to obtain the student's and teachers' label count vectors.
While the label count of local models can be easily obtained (from the class frequency of local datasets), the label count of the global model is not readily available.
We construct $\boldsymbol{\pi}$, the global model's relative label count, as follows.
Let $r_k$ denote the number of rounds that client $k$ has participated in.
For every client $k$ that participates at round $t$, Flashback adds a fraction $\gamma \in (0,1]$ of client $k$ label count ($\boldsymbol{\mu}_k$) to $\boldsymbol{\pi}$, unless $\gamma r_k > 1$, in which case $\boldsymbol{\pi}$ is not updated based on $k$'s label count. The latter case means that client $k$ has participated enough times that its label count is fully accounted for in $\boldsymbol{\pi}$.

Intuitively, the parameter $\gamma$ indicates the rate at which we rely on the global model. When $\gamma$ is set to 1, it implies complete trust in the global model's ability to incorporate the clients' knowledge after just one round of participation. However, expecting such immediate and full assimilation is unrealistic, so we typically set $\gamma < 1$.
The gradual build-up of the global label count plays a vital role in maintaining a balanced distillation in \cref{eq:d_kd_loss} during local updates. This progressive approach mirrors our growing trust in the global model's capabilities. It prevents the risk of assigning excessively high weights too soon, which could otherwise hurt the learning process.

\ifreport
\ifreport
\subsection{Label Count Motivation}
\else
\section{Label Count Motivation}
\fi

\cref{fig:label_count_vs_test_acc_cinic10_fedavg} shows, on the left, the per-class accuracy of a randomly-chosen example client from a \gls{fedavg} training experiment. On the right, the figure shows the corresponding label count at the client. This example suggests that label count can be representative of the model performance. 

\begin{figure*}[h]
    \centering
    \includegraphics[width=0.95\textwidth]{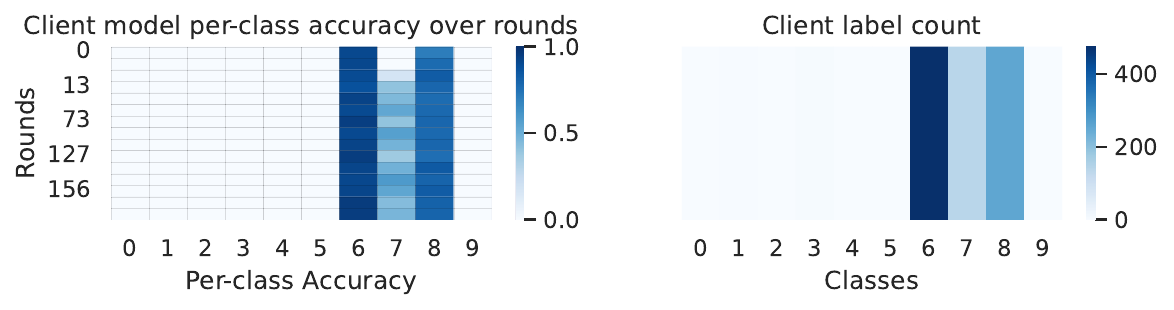}
    \caption{(left) Per-class accuracy of a client model on all the rounds where it participated. (right) Data distribution of that client.}
    \label{fig:label_count_vs_test_acc_cinic10_fedavg}
\end{figure*}

\fi

\section{Experiments \& Results}
\label{sec:eval}

We outline and analyze our experimental findings to investigate whether mitigating forgetting successfully addresses the issues of slow and unstable convergence observed in the initial problem laid out in \cref{sec:intro}. The experimental results stem from three settings: CIFAR10 and CINIC10, where heterogeneous data partitions are created using Dirichlet distribution with $\beta=0.1$ and FEMNIST with 3,432 clients, following the natural heterogeneity of the dataset.
Furthermore, we do an ablation study on the different components of the algorithm.
We use the same neural network architecture that is used in ~\citet{FedNTD-Lee2021-is, FedAvg-McMahan2017-mv}, which is a 2-layer \gls{cnn}.
Summaries of the datasets, partitions, and more details on the experimental setup, as well as additional results, are reported in \cref{sec:exp_details}.

We compare Flashback against several baseline methods, namely:
\begin{inparaenum}[1)]
    \item FedAvg~\citep{FedAvg-McMahan2017-mv},
    \item FedDF~\citep{feddf@lin2020ensemble},
    \item FedNTD~\citep{FedNTD-Lee2021-is},
    \item FedProx~\citep{fedprox@li2020federated},
    \item FedReg~\citep{fedreg@xu2022acceleration},
    \item MOON~\citep{moon@li2021model}.
\end{inparaenum}
It is noteworthy that both FedNTD and FedReg target forgetting in \gls{fl} (discussed further in~\cref{sec:rw}). 
Flashback server distillation is performed until early stopping is triggered on the public validation set (details in \cref{sec:exp_details}). Moreover, Flashback only introduces one additional hyperparameters $\gamma$, which represents how fast trust is built in the global model. We analyze the effect of $\gamma$ later in the section. 
We start by evaluating Flashback performance by showing round-to-accuracy, round forgetting, and the local-global loss over the rounds.

\textbf{Improved round-to-accuracy.}
We evaluate the learning efficiency of Flashback and other baselines by showing the accuracy over rounds in \cref{fig:main_test_acc}. Flashback consistently shows a faster convergence to a high accuracy. Furthermore, we show the number of rounds it takes to reach a target accuracy and fractions of that target accuracy in \cref{table:round-to-accuracy}; we include the result of a central training on the public dataset. Flashback shows a much faster convergence than other baselines. 
This indicates that addressing forgetting on the clients' local update and at the aggregation step does provide training stability and indeed a faster convergence.
\ifreport
\begin{table*}[t!]
\caption{Number of rounds to reach accuracy $A_{x} = A \cdot x$ where $A$ is the target accuracy and $x$ is a fraction of it.}

    \centering
    \begin{tabular}{l|rrr|rrr|rrr}
    \toprule
     & \multicolumn{3}{c|}{CIFAR10, $A=48.2$\%} & \multicolumn{3}{c|}{CINIC10, $A=43.5$\%} & \multicolumn{3}{c}{FEMNIST, $A=69.5$\%} \\
     & $A_{0.5}$ & $A_{0.75}$ & $A_{0.95}$ & $A_{0.5}$ & $A_{0.75}$ & $A_{0.95}$ & $A_{0.5}$ & $A_{0.75}$ & $A_{0.95}$ \\
    \midrule
    FedAvg & 12 & 82 & - & 13 & - & - & 49 & 75 & 138 \\
    FedDF & 7 & 40 & 112 & \textbf{2} & 30 & - & - & - & - \\
    FedNTD & 12 & 41 & 112 & 13 & 46 & - & - & - & - \\
    FedProx & 35 & 93 & - & 13 & - & - & 142 & - & - \\
    FedReg & 35 & 108 & - & 16 & - & - & - & - & - \\
    MOON & 82 & - & - & 124 & - & - & - & - & - \\
    \midrule
    Flashback & \textbf{2} & \textbf{4} & \textbf{10} & 4 & \textbf{5} & \textbf{6} & \textbf{3} & \textbf{5} & \textbf{16} \\
    \bottomrule
    \end{tabular}

\label{table:round-to-accuracy}
\end{table*}
\fi

\textbf{Less round forgetting.}
We show the empirical cumulative distribution function (ECDF) of the round forgetting in \cref{fig:main_round_forgetting}. We see that Flashback successfully reduces round forgetting. Also, FedNTD has less round forgetting than the remaining baselines. In the appendix, we show the round forgetting over the rounds (\cref{fig:round_forgetting}). 

\ifreport
\begin{figure*}[t!]
\centering
\includegraphics[width=0.8\textwidth]{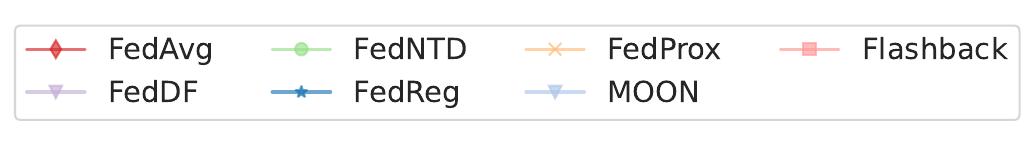}
\begin{subfigure}{0.49\textwidth}
    \includegraphics[width=\textwidth]{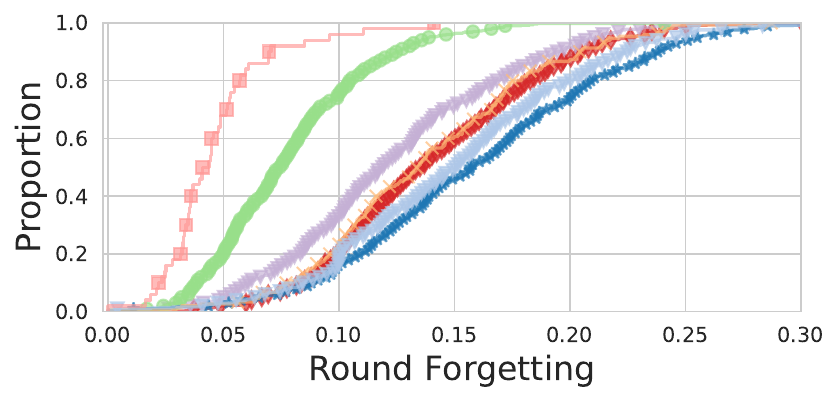}
    \caption{CIFAR10.}
    \label{fig:cifar10_test_round_forgetting_ecdf}
\end{subfigure}
\hfill
\begin{subfigure}{0.49\textwidth}
    \includegraphics[width=\textwidth]{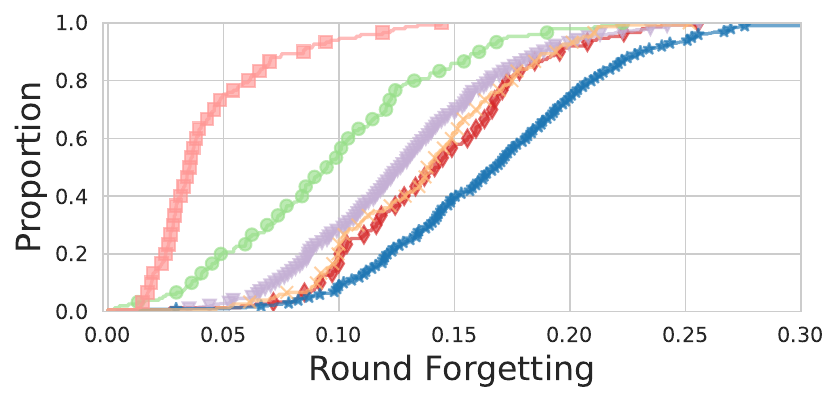}
    \caption{CINIC10.}
    \label{fig:cinic10_test_round_forgetting_ecdf}
\end{subfigure}
\vspace{-1.0em}
\caption{Distribution of round forgetting of Flashback compared to other baselines.}
\label{fig:main_round_forgetting}
\end{figure*}
\fi

\textbf{Minimizing local models divergence.} 
To further understand the effect of Flashback on the training behavior, we show the transition of the mean loss of the local models to the loss of the global model over the rounds in \cref{fig:main_main_local_global_test_loss}. This gives us an insight into the effect of the regularization made by our dynamic distillation. 
We see that the mean loss of the local models of the other baselines always spikes, signifying a divergence of these models from the global training objective, while Flashback has a much more stable loss. This shows that \cref{eq:d_kd_loss} regularizes the local models well so that they do not diverge too much from the global training objective.

In the rest of the section, we delve deeper into Flashback to understand its behavior and validate its performance gains.

\textbf{Dissecting the distillation.}
To show the importance of performing dynamic distillation during clients' updates and at the server's aggregation step, we conduct an experiment where we run Flashback with local distillation only and with server distillation only (c.f. \cref{fig:cifar10_distill_side}). We observe that doing dynamic distillation at either side of the algorithm -- client update and aggregation step -- doesn't address forgetting or gets a similar performance to Flashback. 
\ifreport
\begin{figure*}[t!]
\centering
\includegraphics[width=0.8\textwidth]{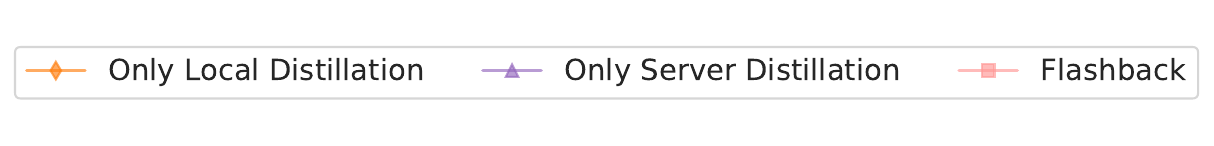}
\begin{subfigure}{0.49\textwidth}
    \includegraphics[width=\textwidth]{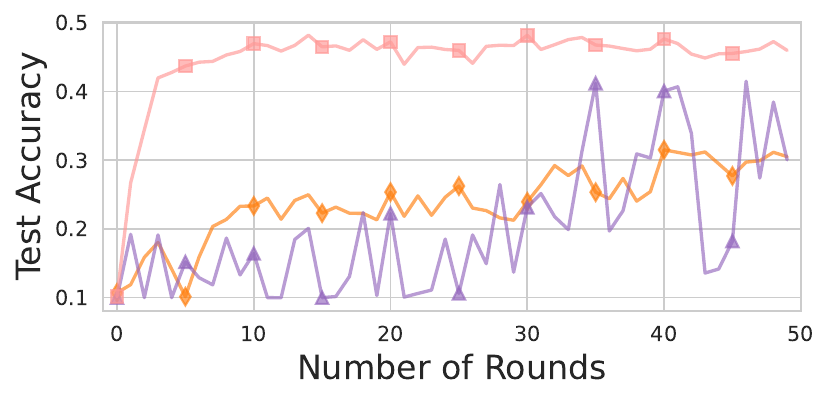}
    \label{fig:cifar10_distill_side_test_acc}
\end{subfigure}
\hfill
\begin{subfigure}{0.49\textwidth}
    \includegraphics[width=\textwidth]{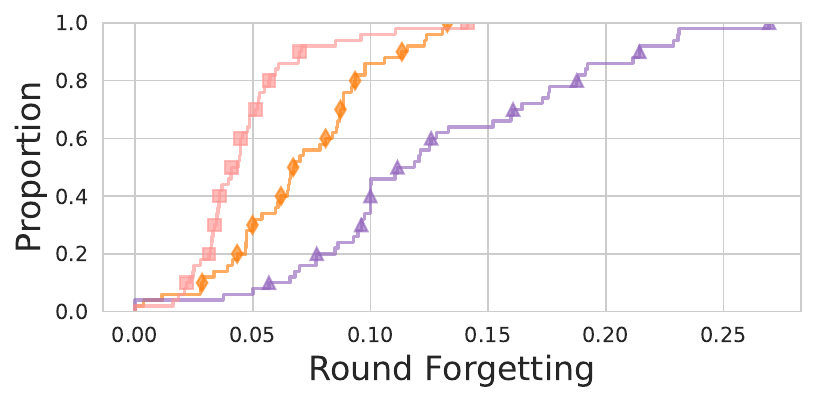}
    \label{fig:cifar10_distill_side/test_round_forgetting_ecdf}
\end{subfigure}
\vspace{-2.5em}
\caption{Performing distillation at only one side of the algorithm (client \& server) on CIFAR10.}
\label{fig:cifar10_distill_side}
\end{figure*}
\fi

To validate the \emph{importance of dynamic distillation at client- and server sides} towards Flashback's performance gains, we create a baseline where we replace the local distillation loss with not-true distillation (NTD) loss~\citet{FedNTD-Lee2021-is}. 
From \cref{fig:flashback_dynamic_kd_vs_NTD}, we observe that this baseline doesn't perform as well as Flashback, and performs similarly to FedNTD.
This indicates that performing the distillation at the server needs well-regularized local models (teachers), which is further supported by our previous experiment contrasting Flashback to single-side distillation (c.f. \cref{fig:cifar10_distill_side}).

\ifreport
\begin{figure*}[t!]
\centering
\includegraphics[width=0.8\textwidth]{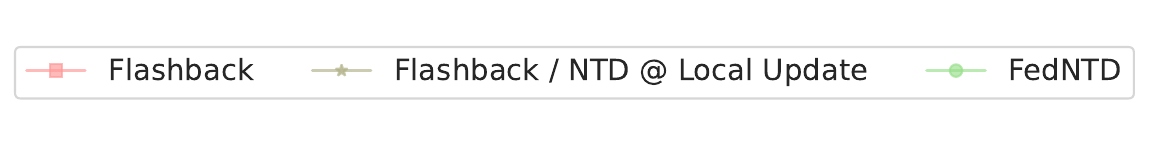}
\begin{subfigure}{0.48\textwidth}
    \includegraphics[width=\textwidth]{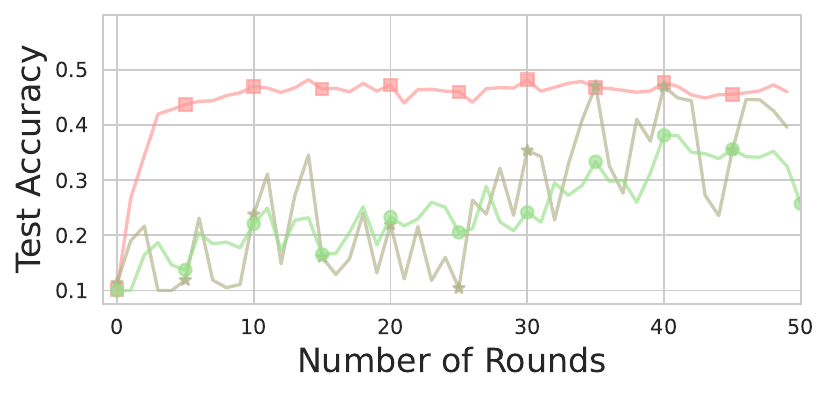}
    \label{fig:flashback_dynamic_kd_vs_NTD/test_acc}
\end{subfigure}
\hfill
\begin{subfigure}{0.48\textwidth}
    \includegraphics[width=\textwidth]{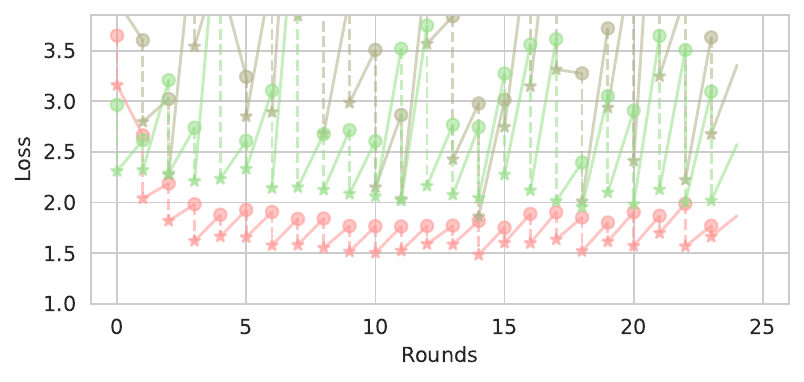}
    \label{fig:flashback_dynamic_kd_vs_NTD/local_global_test_loss}
\end{subfigure}
\vspace{-1.0em}
\caption{Using NTD loss instead of Flashback's dynamic distillation at the local update on CIFAR10.}
\label{fig:flashback_dynamic_kd_vs_NTD}
\end{figure*} 
\fi

\begin{figure}[t!]
\centering
\includegraphics[width=0.95\columnwidth]{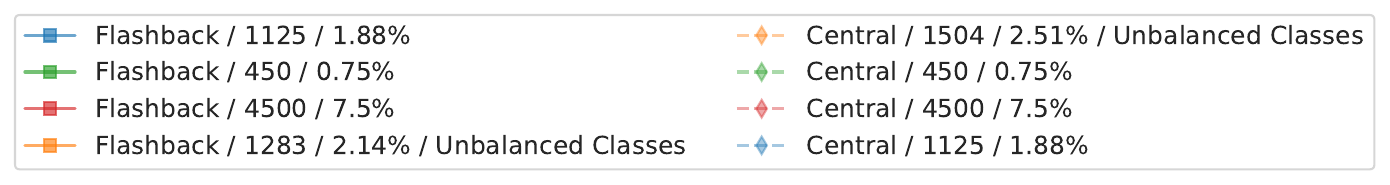}
\centering
\includegraphics[width=0.95\columnwidth]{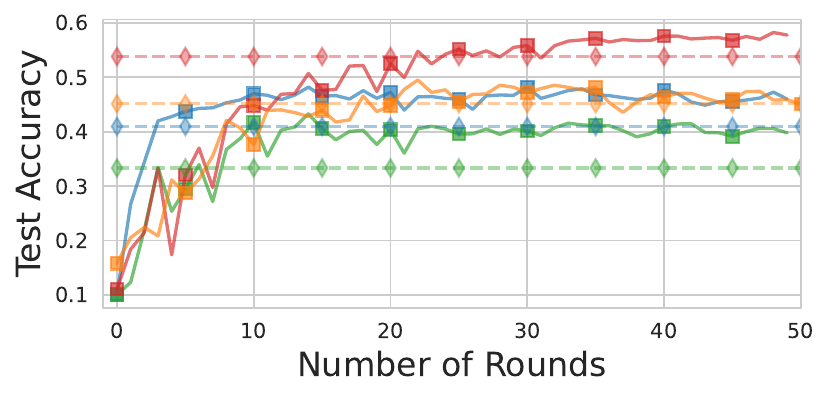}
\vspace{-1.0em}
\caption{Flashback using different public datasets, and the results of central training on public datasets of similar sizes.}
\label{fig:public_exp}
\end{figure}

\textbf{Effect of public dataset size.}
Flashback requires the availability of a public labeled dataset. This may be a limiting assumption in some cases. To study this limitation, we explore a few scenarios for the size of the public dataset in \cref{fig:public_exp}:
\begin{inparaenum}[1)]
    \item 9000 samples ($15$\% of CIFAR10),
    \item 1125 samples ($1.88$\% of CIFAR10),
    \item 1283 samples that have unbalanced class distribution ($2.14$\% of CIFAR10),
    \item 450 samples ($0.75$\% of CIFAR10).
\end{inparaenum}
For all of these scenarios, we train a model centrally on the public dataset.
We find that Flashback can benefit from a large balanced public dataset. Most importantly, Flashback can work well with a small public dataset (1125 samples is the default in all experiments). Furthermore, even if the public dataset has a class imbalance Flashback still performs relatively well. In all of the cases, Flashback always outperforms central training on the public dataset. 
Overall, Flashback requires the availability of a public dataset, however, it does not require a huge amount of data or hard requirements for the class distribution to be very balanced.

\textbf{Training on the public dataset.}
We also experiment to answer the following question: \emph{does the performance improvement of Flashback come from the fact that we train the global model on a public labeled dataset?} To answer this question we create a naive baseline, where we extend FedAvg to fine-tune the global model after the aggregation step at every communication round. From \cref{fig:fedavg_finetuning_repeated}, we see that \gls{fedavg} with fine-tuning quickly reaches a stale model and eventually collapses. We believe this collapse happens due to the local models diverging too much after the local update such that aggregating those models fails. This is evident by the spike of the local model's loss.

\ifreport
\begin{figure*}[t!]
\centering
\begin{subfigure}{0.49\textwidth}
    \includegraphics[width=1\textwidth]{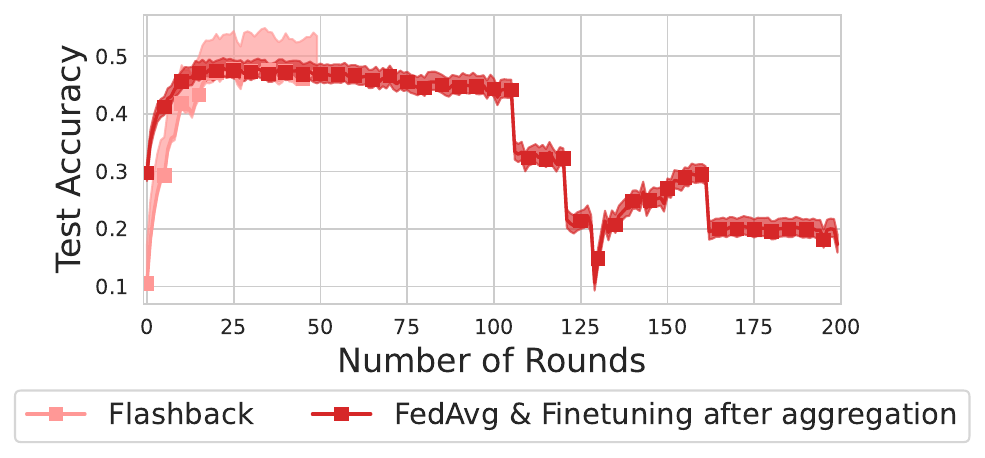}
\end{subfigure}
\hfill
\begin{subfigure}{0.49\textwidth}
    \includegraphics[width=1\textwidth]{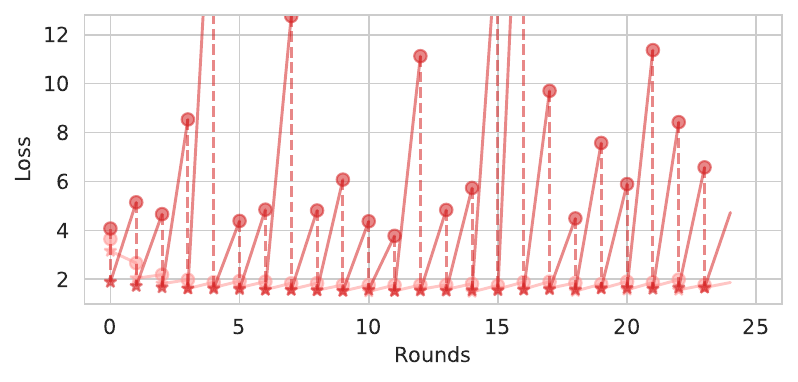}
\end{subfigure}
\vspace{-1.0em}
\caption{Comparing Flashback to FedAvg with fine-tuning: (left) test accuracy over 3 runs on CIFAR10; (right) the  transition of the local models' loss to the global model loss over the rounds.}
\label{fig:fedavg_finetuning_repeated}
\end{figure*}
\fi

\begin{figure}[t!]
\centering
\includegraphics[width=0.95\columnwidth]{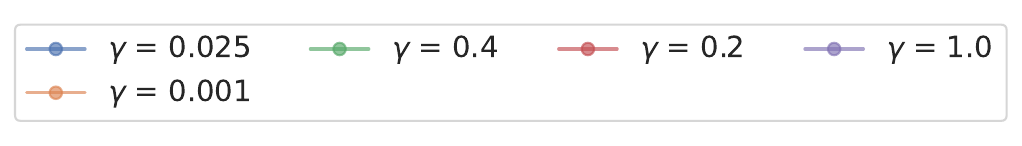}
\centering
\includegraphics[width=0.95\columnwidth]{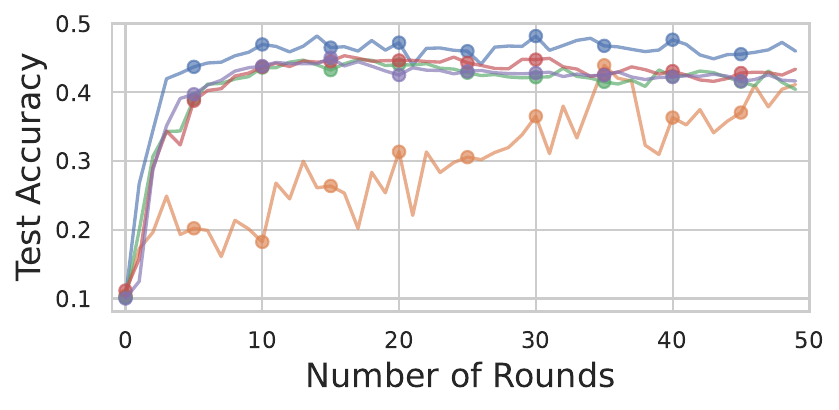}
\vspace{-1.0em}
\caption{Varying the trust of the global model parameter $\gamma$.}
\label{fig:vary_trust_percentage}
\end{figure}

\textbf{The importance of } $\gamma$.
As mentioned, Flashback has a single hyperparameter $\gamma$, which dictates how fast models will trust the global model as a competent teacher. We explore the effect of this hyperparameter in \cref{fig:vary_trust_percentage}. We find that setting this parameter to a larger value leads the learning process to get to a stale solution quickly. This is intuitive since large $\gamma$ leads the global model label count to grow faster, therefore, this dominates the loss term in \cref{eq:d_kd_loss} during both the client- and the server update.
Smaller $\gamma$ gives the best performance because it gives the local models time to learn from their own private data by having a small weight to the distillation term in \cref{eq:d_kd_loss}. However, too small value for $\gamma$ such as $0.001$ slows the training process, since during the local training the distillation term in \cref{eq:d_kd_loss} will be very small in the early rounds. 

\ifreport
\ifreport
\subsection{Additional Results}
\else
\section{Additional Results}
\fi

In \cref{fig:round_forgetting}, we show the round forgetting score computed over the rounds. We see that the baselines have very flaunting round forgetting score.

In \cref{fig:cifar10_distill_side_local_global_test_loss}, we show the transition of the average loss of the local models to the global model loss on the test set. Interestingly, we can see that even though performing local distillation only doesn't have the same performance as Flashback, it does mitigate the local forgetting. That is, we do not see a spike in the loss after the clients perform their local update. 

In \cref{fig:flashback_dynamic_kd_vs_NTD/test_round_forgetting_ecdf}, we see the ECDF of the round forgetting. The Flashback variant with NTD even shows worse forgetting than Flashback and FedNTD. Further showing that just performing distillation at both sides doesn't address the forgetting problem.

\begin{figure*}[t!]
\centering
\includegraphics[width=0.8\textwidth]{figs/cifar10_main/round_forgetting_ecdf_legend.pdf}
\begin{subfigure}{0.48\textwidth}
    \includegraphics[width=\textwidth]{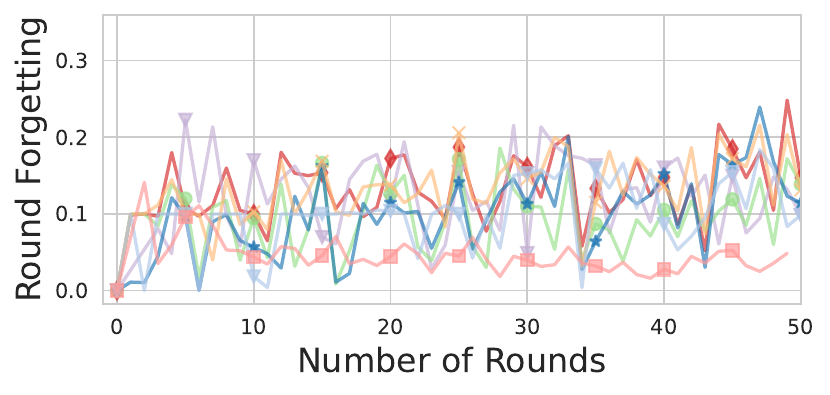}
    \caption{CIFAR10.}
    \label{fig:cifar10_test_round_forgetting}
\end{subfigure}
\hfill
\begin{subfigure}{0.48\textwidth}
    \includegraphics[width=\textwidth]{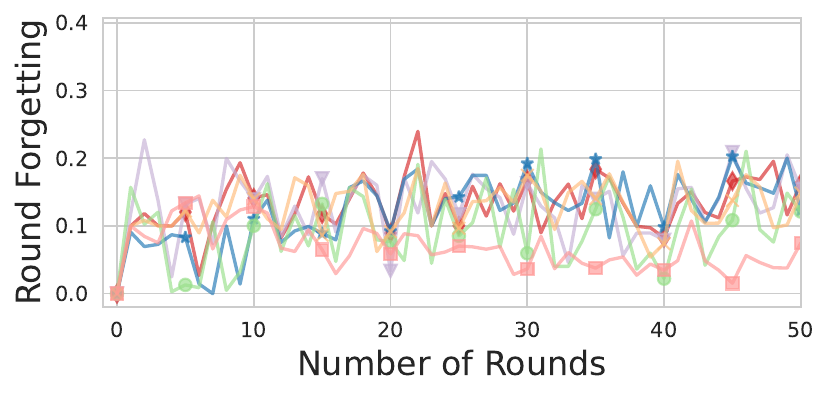}
    \caption{CINIC10.}
    \label{fig:cinic10_test_round_forgetting}
\end{subfigure}
\vspace{-1.0em}
\caption{The round forgetting of Flashback and other baselines over training rounds.}
\label{fig:round_forgetting}
\end{figure*}

\begin{figure}[h]
\centering
\includegraphics[width=0.8\columnwidth]{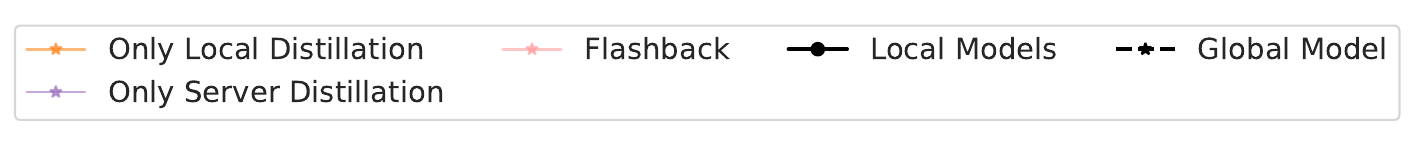}
    \includegraphics[width=\columnwidth]{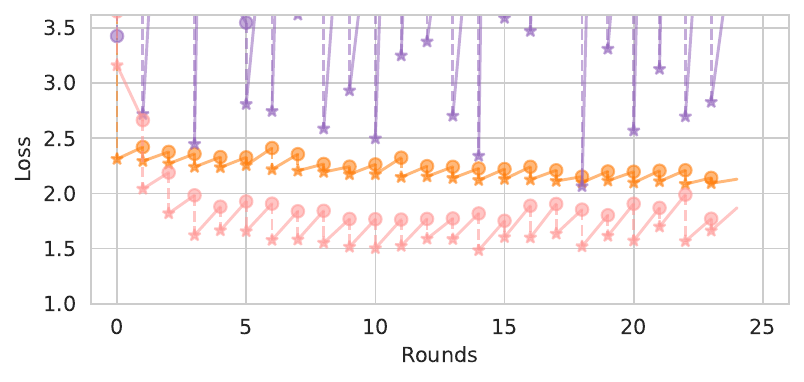}
\vspace{-2.5em}
\caption{Performing distillation at only one side of the algorithm (client \& server) on CIFAR10.}
\label{fig:cifar10_distill_side_local_global_test_loss}
\end{figure}

\begin{figure}[h]
\centering
\includegraphics[width=0.8\columnwidth]{figs/flashback_dynamic_kd_vs_NTD/round_forgetting_ecdf_legend.pdf}
    \includegraphics[width=\columnwidth]{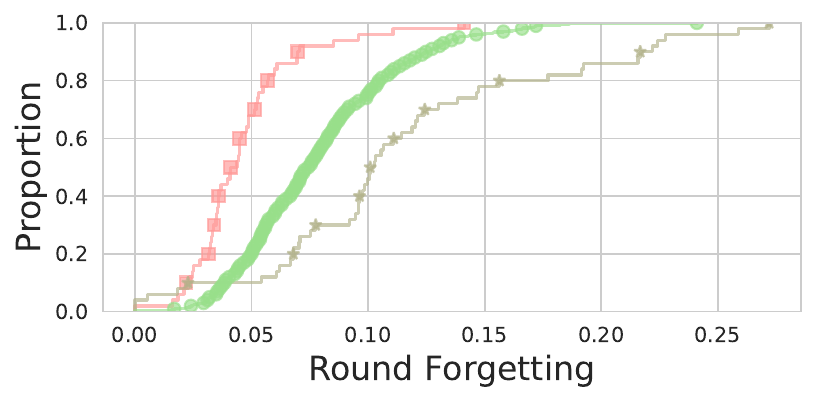}
\vspace{-2.5em}
\caption{Using Not-True Distillation instead of Flashback's dynamic distillation at the local update on CIFAR10.}
\label{fig:flashback_dynamic_kd_vs_NTD/test_round_forgetting_ecdf}
\end{figure}

\fi

\section{Related Work}
\label{sec:rw}

\ifreport
\textbf{Federated learning.} FL is commonly viewed as a ML paradigm wherein a server distributes the training process on a set of decentralized participants that train a shared global model using local datasets that are never shared~\citep{konevcny2015federated, shokri2015privacy, konevcny2016federated, konevcny2017stochastic, fedprox@li2020federated,FedAvg-McMahan2017-mv,fl-open-problems@Kairouz2019-qi}. FL has been used to enhance prediction quality for virtual keyboards among other applications~\citep{Bonawitz19,yang2018applied}. A number of FL frameworks have facilitated research in this area~\citep{caldas2018leaf,pysyft,tff,refl}. 

\textbf{Heterogeneity in FL.} A key challenge in FL systems is uncertainties stemming from learner, system, and data heterogeneity. The non-IID distributions of Learners' data can significantly slow down convergence~\citep{FedAvg-McMahan2017-mv,fl-open-problems@Kairouz2019-qi} and several algorithms are proposed as means of mitigation~\citep{filfl, wang2020tackling,karimireddy2020scaffold,fedprox@li2020federated,Li2021}.

In this section, we review works that targeted the issue of forgetting in \gls{fl}. 
\textbf{Forgetting in \gls{fl}} is an under-studied area that poses significant challenges, leading to slow model convergence and loss of crucial knowledge acquired during the learning process~\citep{forgetting_metric@chaudhry2018riemannian,CFL_Forgetting}. There have been some notable attempts to mitigate the impact of forgetting on the learning process~\citep{FedNTD-Lee2021-is,fedreg@xu2022acceleration}.
\fi

\textbf{FedReg}~\citep{fedreg@xu2022acceleration} addresses the issue of slow convergence in \gls{fl}, asserting it to be a result of forgetting at the local update phase. They demonstrate this by comparing the loss of the global model $w_{t-1}$ on specific client data points with the averaged loss of updated clients' models $\{w_{t,k} \mid k \in \sS_t\}$ on the same data points, highlighting a significant increase in the average loss, indicative of forgetting.
However, in our work, we propose a systematic way of measuring forgetting using a metric designed to capture it. Furthermore, we show that forgetting doesn't only occur in the local update, but it also happens at the aggregation step (\cref{sec:forgetting,fig:global_local_forgetting_cifar_fedavg}).
FedReg proposes to generate fake data that carries the previously attained knowledge. During the local update, Fast Gradient Sign Method~\citep{fgsm@goodfellow2014explaining} is used to generate these data using the global model $w_{t-1}$ and the client data. Then, the loss of the generated data is used to regularize the local update. 
While FedReg employs regularization using synthetic data during local updates, our work, Flashback, leverages dynamic distillation to ensure knowledge retention at both local updates and aggregation steps.

\textbf{FedNTD}~\citep{FedNTD-Lee2021-is} makes a connection between \gls{cl} and \gls{fl}, suggesting that forgetting happens in \gls{fl} as well. Similarly to FedReg, their analysis shows that forgetting happens at the local update, where global knowledge that lies outside of the local distribution of the client is susceptible to forgetting. 
To address this, they propose to use a new variant of distillation~\cref{eq:distillation} named Not-True Distillation (NTD), that masks the ground-truth class logits in the KL divergence as $\mathcal{L}_{\text{KL}}(\vp, \vq) = \sum_{i=c, c \neq y}^C \evp^c \log (\frac{\evp^c}{\evq^c})$, 
where $y$ is the ground-truth class. NTD is used at the local update, while all the other steps in the algorithm remain the same as FedAvg. FedNTD aims to preserve global knowledge during the local update. 

Both FedReg and FedNTD diagnose the issue of forgetting primarily within the realm of local updates, asserting that this stage risks losing valuable global knowledge. Consequently, both works present innovative solutions specifically tailored to counteract this local update forgetting. However, their perspective overlooks a pivotal aspect of the forgetting problem: the occurrence of forgetting during the aggregation step. As we delve into in~\cref{sec:forgetting}, this oversight in recognizing and addressing forgetting during aggregation has repercussions on the later local updates. In contrast, Flashback takes a holistic approach, targeting forgetting comprehensively across both the local updates and the aggregation phase, leading to faster convergence.

\section{Conclusion}
We explored the phenomenon of forgetting in \gls{fl}. Our investigation revealed that forgetting occurs during both local and global update phases of FL algorithms.
We presented Flashback, a novel \gls{fl} algorithm explicitly designed to counteract forgetting by employing dynamic knowledge distillation. Our approach leverages data label counts as a proxy for knowledge, ensuring a more targeted and effective forgetting mitigation.
Our empirical results showed Flashback's efficacy in mitigating round forgetting, thereby supporting the hypothesis that the observed slow and unstable convergence in FL algorithms is closely linked to forgetting. This result underlines the importance of addressing forgetting, paving the way for the advancement of more robust and efficient FL algorithms.

\bibliographystyle{icml2024}
\bibliography{ref}

\begin{thebibliography}{33}
\providecommand{\natexlab}[1]{#1}
\providecommand{\url}[1]{\texttt{#1}}
\expandafter\ifx\csname urlstyle\endcsname\relax
  \providecommand{\doi}[1]{doi: #1}\else
  \providecommand{\doi}{doi: \begingroup \urlstyle{rm}\Url}\fi

\bibitem[Abdelmoniem et~al.(2023)Abdelmoniem, Sahu, Canini, and Fahmy]{refl}
Abdelmoniem, A.~M., Sahu, A.~N., Canini, M., and Fahmy, S.~A.
\newblock {REFL: Resource-Efficient Federated Learning}.
\newblock In \emph{ACM EuroSys}, 2023.

\bibitem[Bonawitz et~al.(2019)Bonawitz, Eichner, Grieskamp, Huba, Ingerman, Ivanov, Kiddon, Kone\v{c}n\'{y}, Mazzocchi, McMahan, Van~Overveldt, Petrou, Ramage, and Roselander]{Bonawitz19}
Bonawitz, K., Eichner, H., Grieskamp, W., Huba, D., Ingerman, A., Ivanov, V., Kiddon, C., Kone\v{c}n\'{y}, J., Mazzocchi, S., McMahan, B., Van~Overveldt, T., Petrou, D., Ramage, D., and Roselander, J.
\newblock {Towards Federated Learning at Scale: System Design}.
\newblock In \emph{MLSys}, 2019.

\bibitem[Bucilu\v{a} et~al.(2006)Bucilu\v{a}, Caruana, and Niculescu-Mizil]{bucilua2006model}
Bucilu\v{a}, C., Caruana, R., and Niculescu-Mizil, A.
\newblock Model compression.
\newblock In \emph{KDD}, 2006.

\bibitem[Caldas et~al.(2019)Caldas, Duddu, Wu, Li, Konečný, McMahan, Smith, and Talwalkar]{caldas2018leaf}
Caldas, S., Duddu, S. M.~K., Wu, P., Li, T., Konečný, J., McMahan, H.~B., Smith, V., and Talwalkar, A.
\newblock {LEAF: A Benchmark for Federated Settings}.
\newblock In \emph{Workshop on Federated Learning for Data Privacy and Confidentiality}, 2019.

\bibitem[Chaudhry et~al.(2018)Chaudhry, Dokania, Ajanthan, and Torr]{forgetting_metric@chaudhry2018riemannian}
Chaudhry, A., Dokania, P.~K., Ajanthan, T., and Torr, P.~H.
\newblock Riemannian walk for incremental learning: Understanding forgetting and intransigence.
\newblock In \emph{Proceedings of the European conference on computer vision (ECCV)}, 2018.

\bibitem[Darlow et~al.(2018)Darlow, Crowley, Antoniou, and Storkey]{cinic10@darlow2018cinic}
Darlow, L.~N., Crowley, E.~J., Antoniou, A., and Storkey, A.~J.
\newblock Cinic-10 is not imagenet or cifar-10.
\newblock \emph{arXiv preprint arXiv:1810.03505}, 2018.

\bibitem[De~Lange et~al.(2021)De~Lange, Aljundi, Masana, Parisot, Jia, Leonardis, Slabaugh, and Tuytelaars]{f_cl2021continual}
De~Lange, M., Aljundi, R., Masana, M., Parisot, S., Jia, X., Leonardis, A., Slabaugh, G., and Tuytelaars, T.
\newblock A continual learning survey: Defying forgetting in classification tasks.
\newblock \emph{IEEE transactions on pattern analysis and machine intelligence}, 2021.

\bibitem[Dupuy et~al.(2023)Dupuy, Majmudar, Wang, Roosta, Gupta, Chung, Ding, and Avestimehr]{CFL_Forgetting}
Dupuy, C., Majmudar, J., Wang, J., Roosta, T.~G., Gupta, R., Chung, C., Ding, J., and Avestimehr, S.
\newblock Quantifying catastrophic forgetting in continual federated learning.
\newblock In \emph{IEEE International Conference on Acoustics, Speech and Signal Processing (ICASSP)}, 2023.

\bibitem[Fourati et~al.(2023)Fourati, Kharrat, Aggarwal, Alouini, and Canini]{filfl}
Fourati, F., Kharrat, S., Aggarwal, V., Alouini, M.-S., and Canini, M.
\newblock {FilFL: Accelerating Federated Learning via Client Filtering}, 2023.
\newblock URL \url{https://arxiv.org/abs/2302.06599}.

\bibitem[Goodfellow et~al.(2014)Goodfellow, Shlens, and Szegedy]{fgsm@goodfellow2014explaining}
Goodfellow, I.~J., Shlens, J., and Szegedy, C.
\newblock Explaining and harnessing adversarial examples.
\newblock \emph{arXiv preprint arXiv:1412.6572}, 2014.

\bibitem[Hinton et~al.(2015)Hinton, Vinyals, and Dean]{hinton2015distilling}
Hinton, G., Vinyals, O., and Dean, J.
\newblock Distilling the knowledge in a neural network, 2015.

\bibitem[Jeong et~al.(2018)Jeong, Oh, Kim, Park, Bennis, and Kim]{FD-Jeong2018-ew}
Jeong, E., Oh, S., Kim, H., Park, J., Bennis, M., and Kim, S.-L.
\newblock {Communication-Efficient} {On-Device} machine learning: Federated distillation and augmentation under {Non-IID} private data.
\newblock 2018.

\bibitem[Kairouz et~al.(2019)Kairouz, Brendan~McMahan, Avent, Bellet, Bennis, Bhagoji, Bonawitz, Charles, Cormode, Cummings, D'Oliveira, Eichner, El~Rouayheb, Evans, Gardner, Garrett, Gasc{\'o}n, Ghazi, Gibbons, Gruteser, Harchaoui, He, He, Huo, Hutchinson, Hsu, Jaggi, Javidi, Joshi, Khodak, Kone{\v c}n{\'y}, Korolova, Koushanfar, Koyejo, Lepoint, Liu, Mittal, Mohri, Nock, {\"O}zg{\"u}r, Pagh, Raykova, Qi, Ramage, Raskar, Song, Song, Stich, Sun, Suresh, Tram{\`e}r, Vepakomma, Wang, Xiong, Xu, Yang, Yu, Yu, and Zhao]{fl-open-problems@Kairouz2019-qi}
Kairouz, P., Brendan~McMahan, H., Avent, B., Bellet, A., Bennis, M., Bhagoji, A.~N., Bonawitz, K., Charles, Z., Cormode, G., Cummings, R., D'Oliveira, R. G.~L., Eichner, H., El~Rouayheb, S., Evans, D., Gardner, J., Garrett, Z., Gasc{\'o}n, A., Ghazi, B., Gibbons, P.~B., Gruteser, M., Harchaoui, Z., He, C., He, L., Huo, Z., Hutchinson, B., Hsu, J., Jaggi, M., Javidi, T., Joshi, G., Khodak, M., Kone{\v c}n{\'y}, J., Korolova, A., Koushanfar, F., Koyejo, S., Lepoint, T., Liu, Y., Mittal, P., Mohri, M., Nock, R., {\"O}zg{\"u}r, A., Pagh, R., Raykova, M., Qi, H., Ramage, D., Raskar, R., Song, D., Song, W., Stich, S.~U., Sun, Z., Suresh, A.~T., Tram{\`e}r, F., Vepakomma, P., Wang, J., Xiong, L., Xu, Z., Yang, Q., Yu, F.~X., Yu, H., and Zhao, S.
\newblock Advances and open problems in federated learning.
\newblock \emph{arXiv 1912.04977}, 2019.

\bibitem[Karimireddy et~al.(2020)Karimireddy, Kale, Mohri, Reddi, Stich, and Suresh]{karimireddy2020scaffold}
Karimireddy, S.~P., Kale, S., Mohri, M., Reddi, S., Stich, S., and Suresh, A.~T.
\newblock Scaffold: Stochastic controlled averaging for federated learning.
\newblock In \emph{International conference on machine learning}, pp.\  5132--5143, 2020.

\bibitem[Kone{\v{c}}n{\`y}(2017)]{konevcny2017stochastic}
Kone{\v{c}}n{\`y}, J.
\newblock Stochastic, distributed and federated optimization for machine learning.
\newblock \emph{arXiv preprint arXiv:1707.01155}, 2017.

\bibitem[Kone{\v{c}}n{\`y} et~al.(2015)Kone{\v{c}}n{\`y}, McMahan, and Ramage]{konevcny2015federated}
Kone{\v{c}}n{\`y}, J., McMahan, B., and Ramage, D.
\newblock Federated optimization: Distributed optimization beyond the datacenter.
\newblock \emph{arXiv preprint arXiv:1511.03575}, 2015.

\bibitem[Kone{\v{c}}n{\`y} et~al.(2016)Kone{\v{c}}n{\`y}, McMahan, Yu, Richt{\'a}rik, Suresh, and Bacon]{konevcny2016federated}
Kone{\v{c}}n{\`y}, J., McMahan, H.~B., Yu, F.~X., Richt{\'a}rik, P., Suresh, A.~T., and Bacon, D.
\newblock Federated learning: Strategies for improving communication efficiency.
\newblock \emph{arXiv preprint arXiv:1610.05492}, 2016.

\bibitem[Krizhevsky(2009)]{cifar10@krizhevsky2009learning}
Krizhevsky, A.
\newblock Learning multiple layers of features from tiny images.
\newblock Technical report, University of Toronto, 2009.

\bibitem[Lee et~al.(2021)Lee, Jeong, Shin, Bae, and Yun]{FedNTD-Lee2021-is}
Lee, G., Jeong, M., Shin, Y., Bae, S., and Yun, S.-Y.
\newblock Preservation of the global knowledge by {Not-True} distillation in federated learning.
\newblock 2021.

\bibitem[Li et~al.(2021{\natexlab{a}})Li, Duan, Liu, Zhang, Ren, Chen, Tan, and Wang]{Li2021}
Li, L., Duan, M., Liu, D., Zhang, Y., Ren, A., Chen, X., Tan, Y., and Wang, C.
\newblock {FedSAE: A Novel Self-Adaptive Federated Learning Framework in Heterogeneous Systems}.
\newblock In \emph{IJCNN}, 2021{\natexlab{a}}.

\bibitem[Li et~al.(2021{\natexlab{b}})Li, He, and Song]{moon@li2021model}
Li, Q., He, B., and Song, D.
\newblock Model-contrastive federated learning.
\newblock In \emph{Conference on computer vision and pattern recognition}, 2021{\natexlab{b}}.

\bibitem[Li et~al.(2020)Li, Sahu, Zaheer, Sanjabi, Talwalkar, and Smith]{fedprox@li2020federated}
Li, T., Sahu, A.~K., Zaheer, M., Sanjabi, M., Talwalkar, A., and Smith, V.
\newblock Federated optimization in heterogeneous networks.
\newblock \emph{Proceedings of Machine learning and systems}, 2020.

\bibitem[Lin et~al.(2020)Lin, Kong, Stich, and Jaggi]{feddf@lin2020ensemble}
Lin, T., Kong, L., Stich, S.~U., and Jaggi, M.
\newblock Ensemble distillation for robust model fusion in federated learning.
\newblock \emph{Advances in Neural Information Processing Systems}, 2020.

\bibitem[McMahan et~al.(2017)McMahan, Moore, Ramage, Hampson, and Arcas]{FedAvg-McMahan2017-mv}
McMahan, B., Moore, E., Ramage, D., Hampson, S., and Arcas, B. A.~y.
\newblock {{Communication-Efficient} Learning of Deep Networks from Decentralized Data}.
\newblock In Singh, A. and Zhu, J. (eds.), \emph{AISTATS}, 2017.

\bibitem[OpenMined(2020)]{pysyft}
OpenMined.
\newblock Syft + grid provides secure and private deep learning in python, 2020.
\newblock URL \url{https://github.com/OpenMined/PySyft}.

\bibitem[Parisi et~al.(2019)Parisi, Kemker, Part, Kanan, and Wermter]{forgetting_in_cl@parisi2019continual}
Parisi, G.~I., Kemker, R., Part, J.~L., Kanan, C., and Wermter, S.
\newblock Continual lifelong learning with neural networks: A review.
\newblock \emph{Neural networks}, 2019.

\bibitem[Schmidhuber(1991)]{schmidhuber1991neural}
Schmidhuber, J.
\newblock \emph{Neural sequence chunkers}.
\newblock Inst. f{\"u}r Informatik, 1991.

\bibitem[Shokri \& Shmatikov(2015)Shokri and Shmatikov]{shokri2015privacy}
Shokri, R. and Shmatikov, V.
\newblock Privacy-preserving deep learning.
\newblock In \emph{Proceedings of the 22nd ACM SIGSAC conference on computer and communications security}, pp.\  1310--1321, 2015.

\bibitem[tensorflow.org(2020)]{tff}
tensorflow.org.
\newblock Tensorflow federated: Machine learning on decentralized data, 2020.
\newblock URL \url{https://www.tensorflow.org/federated}.

\bibitem[Wang et~al.(2020{\natexlab{a}})Wang, Yurochkin, Sun, Papailiopoulos, and Khazaeni]{fedma@wang2020federated}
Wang, H., Yurochkin, M., Sun, Y., Papailiopoulos, D., and Khazaeni, Y.
\newblock Federated learning with matched averaging.
\newblock \emph{arXiv preprint arXiv:2002.06440}, 2020{\natexlab{a}}.

\bibitem[Wang et~al.(2020{\natexlab{b}})Wang, Liu, Liang, Joshi, and Poor]{wang2020tackling}
Wang, J., Liu, Q., Liang, H., Joshi, G., and Poor, H.~V.
\newblock Tackling the objective inconsistency problem in heterogeneous federated optimization.
\newblock In \emph{NeurIPS}, 2020{\natexlab{b}}.

\bibitem[Xu et~al.(2022)Xu, Hong, Huang, and Jiang]{fedreg@xu2022acceleration}
Xu, C., Hong, Z., Huang, M., and Jiang, T.
\newblock Acceleration of federated learning with alleviated forgetting in local training.
\newblock \emph{arXiv preprint arXiv:2203.02645}, 2022.

\bibitem[Yang et~al.(2018)Yang, Andrew, Eichner, Sun, Li, Kong, Ramage, and Beaufays]{yang2018applied}
Yang, T., Andrew, G., Eichner, H., Sun, H., Li, W., Kong, N., Ramage, D., and Beaufays, F.
\newblock {Applied Federated Learning: Improving Google Keyboard Query Suggestions}, 2018.

\end{thebibliography}

\clearpage
\appendix

\ifreport

\else

\fi
\ifreport

\else

\fi

\section{Experiments Details}
\label{sec:exp_details}

\subsection{Datasets}
\label{sec:datasets}

In this section, we provide an overview of the datasets used, the data split, and the specific experimental setups. For each dataset, we perform two sets of experiments to analyze the effects of data heterogeneity on the algorithms' performance. The datasets used are CIFAR10, CINIC10, and FEMNIST.

\textbf{CIFAR10} ~\citep{cifar10@krizhevsky2009learning}. A famous vision dataset that includes 50k training images and 10k testing images. We emulate a realistic, heterogeneous data distribution by using a Dirichlet distribution with  $\beta=0.1$. A $\beta$ value of $0.1$ is chosen to simulate a more heterogeneous, and challenging data distribution. A $2.5$\% random sample of the training set creates a public dataset, further divided into training and validation sets. This yields a very small public training dataset with size of $1.88$\%.  The remaining $97.5$\% is distributed among 100 clients, with each client's data being split into training (90\%) and validation (10\%) subsets.

\textbf{CINIC10}~\citep{cinic10@darlow2018cinic}. A drop-in replacement of CIFAR10, this dataset comprises 90k training, 90k validation, and 90k test images. We merge the training and validation sets and adopt a similar approach as with CIFAR10, taking out $2.5$\% of the data to be the public dataset, similar to the CIFAR10 case this $2.5$\% is further divided into training set and validation set. Then employing Dirichlet distribution with $\beta$ value of 0.1 to split the $97.5$\% remaining data into 200 clients, with each client’s data further divided into training (90\%) and validation (10\%) sets. 

\begin{figure}[h]
\begin{subfigure}{\columnwidth}
    \includegraphics[width=\columnwidth]{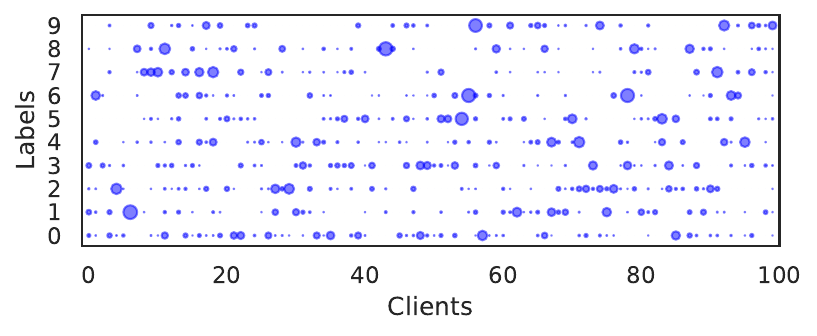}
    \caption{CIFAR10, Dir$(\beta=0.1)$}
    \label{fig:CIFARDataModule_0.1_clients_data_dist}
\end{subfigure}
\vfill
\begin{subfigure}{\columnwidth}
    \includegraphics[width=\columnwidth]{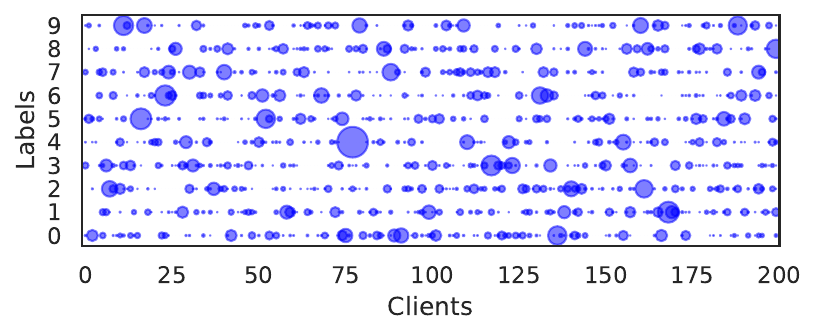}
    \caption{CINIC10, Dir$(\beta=0.1)$}
    \label{fig:CINIC10DataModule_0.1_clients_data_dist}
\end{subfigure}
\vfill
\begin{subfigure}{\columnwidth}
    \includegraphics[width=\columnwidth]{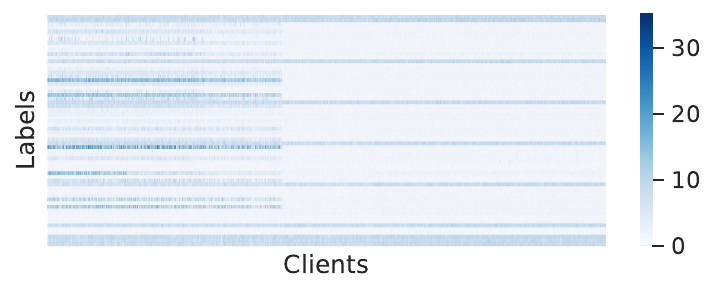}
    \caption{FEMNIST with 3432 clients}
    \label{fig:FEMNISTDataModule_3432_clients_data_dist}
\end{subfigure}
\caption{Clients data distribution. The x-axis is the clients and the y-axis is the labels.}
\label{fig:clients_data_dist}
\end{figure}

\textbf{FEMNIST}~\citep{caldas2018leaf}. This federated learning dataset is based on extended MNIST with natural heterogeneity, where each writer is considered a client. From the 3597 total writers, those with less than 50 samples are excluded. We randomly selected 150 writers to form a public dataset. The remaining 3432 writers' data is divided into train (approx. 70\%), validation (approx. 15\%), and test (approx. 15\%) sets. The collective test sets from all writers form the overall test set. At every round, 32 clients are randomly selected for participation. 

For CIFAR10 and CINIC10, we chose $\beta$ values of 0.1 and client participation value of 10. While for FEMNIST we have 3432 clients (writers) with client participation vale of 32
In all cases, the training data distribution among clients is illustrated in \cref{fig:clients_data_dist}.

\subsection{Baselines \& Hyperparameters}

We evaluate the following algorithms as baselines:
\begin{inparaenum}
    \item FedAvg~\citep{FedAvg-McMahan2017-mv};
    \item FedDF~\citep{feddf@lin2020ensemble};
    \item FedNTD~\citep{FedNTD-Lee2021-is};
    \item FedProx~\citep{fedprox@li2020federated};
    \item FedReg~\citep{fedreg@xu2022acceleration}; and
    \item MOON~\citep{moon@li2021model}
\end{inparaenum}.
Both FedNTD and FedReg target forgetting in \gls{fl} (discussed in~\cref{sec:rw}). We use the same neural network architecture that is used in ~\citet{FedNTD-Lee2021-is, FedAvg-McMahan2017-mv}, which is a 2-layer \gls{cnn}. Note that for MOON ~\citep{moon@li2021model} we add an additional layers to the model as described in their source code for the projection head. 
Moreover, for the optimizer, learning rate, and model we follow~\citet{FedNTD-Lee2021-is, FedAvg-McMahan2017-mv}, and when a baseline has different hyperparameters we use their proposed values. For example, in FedDF the number of local epochs is set to 40, while in the other baselines and Flashback, it is set to 5 epochs. As for Flashback hyperparameters, during the server distillation, we train until early stopping gets triggered using the validation set; we set the label count fraction $\gamma = 0.025$ for CIFAR10, i.e., we add $2.5$\% of the client label count each time it participates, while we set $\gamma = 0.1$ for CINIC10 and FEMNIST. As for distillation-specific hyperparameters, we have one fewer hyperparameter since $\boldsymbol{\alpha}$ is computed automatically, and for temperature, we use the standard $T=3$.

\ifreport

\else

\fi

\end{document}